\documentclass[journal, 9pt]{IEEEtran}
\usepackage[pdftex]{graphicx}

\usepackage[cmex10]{amsmath}
\usepackage{array}
\usepackage{enumerate}
\usepackage{multicol}
\usepackage{multirow}
\usepackage{wrapfig}
\usepackage{subfig}
\usepackage{url}
\usepackage[normalem]{ulem}
\usepackage{bm} 
\usepackage{siunitx}
\usepackage{fixmath} 
\usepackage{algorithm} 
\usepackage{algorithmic}
\usepackage{arydshln}
\usepackage{multirow}
\usepackage{verbatim}
\usepackage{rotating}
\usepackage{textcomp}
\usepackage{amsmath}
\usepackage{amssymb}
\usepackage{booktabs}
\usepackage{cite}
\usepackage{threeparttable}
\usepackage{array}
\usepackage[usenames,dvipsnames]{color, colortbl}
\definecolor{Gray}{gray}{0.9}
\definecolor{LightCyan}{rgb}{0.88,1,1}
\definecolor{ulascolor}{rgb}{0.6, 0.3, 0.2}
\definecolor{dark_purple}{rgb}{0.4, 0.0, 0.4}
\usepackage{amsfonts}
\usepackage{mathtools}
\DeclarePairedDelimiter{\abs}{\lvert}{\rvert}
\newcommand\VRule[1][\arrayrulewidth]{\vrule width #1}

\begin{document}
\title{Instance-Level Microtubule Tracking}

\author{Samira~Masoudi$^{1,2}$,~\IEEEmembership{Student Member, ~IEEE,}
        Afsaneh~Razi$^{1}$,~\IEEEmembership{}
        Cameron~H.G.~Wright$^{2}$,~\IEEEmembership{Senior Member,~IEEE,}
        Jesse~C.~Gatlin$^{2}$,~\IEEEmembership{}
        Ulas~Bagci$^{1}$,~\IEEEmembership{Senior Member,~IEEE}

\thanks{$^{1}$ Masoudi, Razi, and Bagci are with University of Central Florida, Orlando, 32816 FL.}
\thanks{$^{2}$Masoudi, Wright, and Gatlin are with University of Wyoming, Laramie, 82071 WY.}}
\markboth{IEEE Transactions on Medical Imaging,~Vol.~xxx, No.~xxx, Month~2019}%
{Masoudi et al.: Instance-Level MT Tracking}

\maketitle
\begin{abstract}
We propose a new method of instance-level microtubule (MT) tracking in time-lapse image series using recurrent attention. Our novel deep learning algorithm segments individual MTs at each frame. Segmentation results from successive frames are used to assign correspondences among MTs. This ultimately generates a distinct path trajectory for each MT through the frames. Based on these trajectories, we estimate MT velocities. To validate our proposed technique, we conduct experiments using real and simulated data. We use statistics derived from real time-lapse series of MT gliding assays to simulate realistic MT time-lapse image series in our simulated data. This data set is employed as pre-training and hyperparameter optimization for our network before training on the real data. Our experimental results show that the proposed \textcolor{black}{supervised} learning algorithm improves the precision for MT instance velocity estimation drastically to 71.3\%  from the baseline result (29.3\%). We also demonstrate how the inclusion of temporal information into our deep network can reduce the false negative rates from 67.8\% (baseline) down to 28.7\% (proposed). Our findings in this work are expected to help biologists \textcolor{black}{characterize the spatial arrangement of MTs, specifically the effects of MT-MT interactions}.
\end{abstract}


\IEEEpeerreviewmaketitle

 \section{Introduction}
\IEEEPARstart{M}{icrotubules} (MTs) are cytoskeletal polymers within eukaryotic cells composed of individual $\alpha$- and $\beta$-tubulin subunits with head-to-tail arrangement.  The inherent asymmetry of the heterodimeric subunits produces polar filaments with two distinct ends (one termed ``plus" for the dynamic end and the other ``minus" for the more stable end)~\cite{Applegate_thsis,MP_Kreis}. The polar structure of MTs and their highly regulated growth dynamics make them good candidates for many \textcolor{black}{tasks vital to the maintenance of cell homeostasis including intracellular transport, cell migration, asymmetric polarization, and cell division~\cite{MP_Kreis}.} MTs are the primary components of the mitotic spindle which is assembled by the cell to segregate sister chromatids during cell division. \textcolor{black}{Considering their fundamental roles in myriad cellular processes}, it is not surprising that perturbations of MT function can lead to diseases ranging from cancer to neurodegenerative disorders~\cite{pprezi}. Therefore, a quantitative analysis of MTs  is important for understanding the mechanistic underpinnings of many diseases at the molecular level. 

\textcolor{black}{In this study, we characterize two distinct types of MTs' dynamics}: 
\begin{enumerate}
    \item \textcolor{black}{\textit{Individual MT growth dynamics}}: each single MT (with any mobility status: mobile, immobile, others) undergoes several stochastic transitions between growth (subunit attachment), pause, and shrinkage (subunit detachment) at its plus end~\cite{mitchison1984dynamic}. This is called \textit{dynamic instability}.
    \item \textcolor{black}{\textit{Interactions between MT instances}: motor-dependent movement causes MTs to interact with} their surroundings and particularly with other MTs. This interaction occurs through direct contact and/or by crosslinking via specific motor and non-motor proteins known as \textit{microtubule associated proteins} (MAPs)~\cite{MP_Kreis,Pampaloni2008}.
\end{enumerate}
While various investigations have focused on dynamic instability by tracking MT plus-ends only~\cite{MP_Kreis}, the second dynamic type, i.e., changes in \textcolor{black}{MT behaviour}, due to interactions with motors, other proteins and MTs, is still an area that requires additional investigation. \textcolor{black}{MTs are arranged in space by motor-dependent crosslinking~\cite{mcintosh1969anaphase}. The resultant movement is thought to be dependent upon the motor type and density as well as the number and nature of static non-motile MAPs~\cite{mcintosh1969anaphase}}. Due to \textit{sliding-filament} mechanisms, the combined actions of these active and static MAPs, spatially organize MTs relative to each other~\cite{Walczak2010,mcintosh1969anaphase}. 
\textcolor{black}{Intracellular complexity often precludes informative \textit{in-vivo} investigation on such sliding filament mechanisms. Experimentalists can circumvent this limitation by employing reductionist \textit{in-vitro} approaches termed as MT-gliding assays~\cite{mcintosh1969anaphase}. In these assays, MTs are labeled and tracked as they are moved along a coverslip surface by surface-bound MT-dependent motors~\cite{mcintosh1969anaphase}. Although this class of assays has been used extensively~\cite{mcintosh1969anaphase}, the inherent utility of the approach is limited by a general lack of available, objective, and automated tracking methods.  Here we describe the development of a gliding assay analysis method that minimizes subjectivity by applying recurrent attention to identify and segment MTs.} 

\textcolor{black}{To generate novel data for our analysis, we perform similar assays using MTs assembled} in cell-free extracts derived from Xenopus laevis eggs~\cite{desai1999kin}. In these assays, MTs are spiked with fluorescently labeled MAPs ~\cite{mooney2017tau}. Besides, endogenous cytoplasmic motors in the extract bind to the coverslip surface nonspecifically to power the MT gliding. The extract also contains a large complement of non-motor MAPs \textcolor{black}{that are thought to decorate MTs along their lengths and potentially affect MT-MT interactions via binding and/or crosslinking. The depth of the flow chamber used in these studies} is $\sim$50 to 60 times greater than 25nm diameter of MTs, providing sufficient space to enable multiple MTs to freely slide over each other. 
\textcolor{black}{Total internal reflection fluorescence (TIRF) microscopy is used to visualize MT movements and dynamics. Time-lapse image sequences are recorded from TIRF microscopy for our analyses.}
\begin{figure*}[t]
\centering
\includegraphics[width=.70\textwidth]{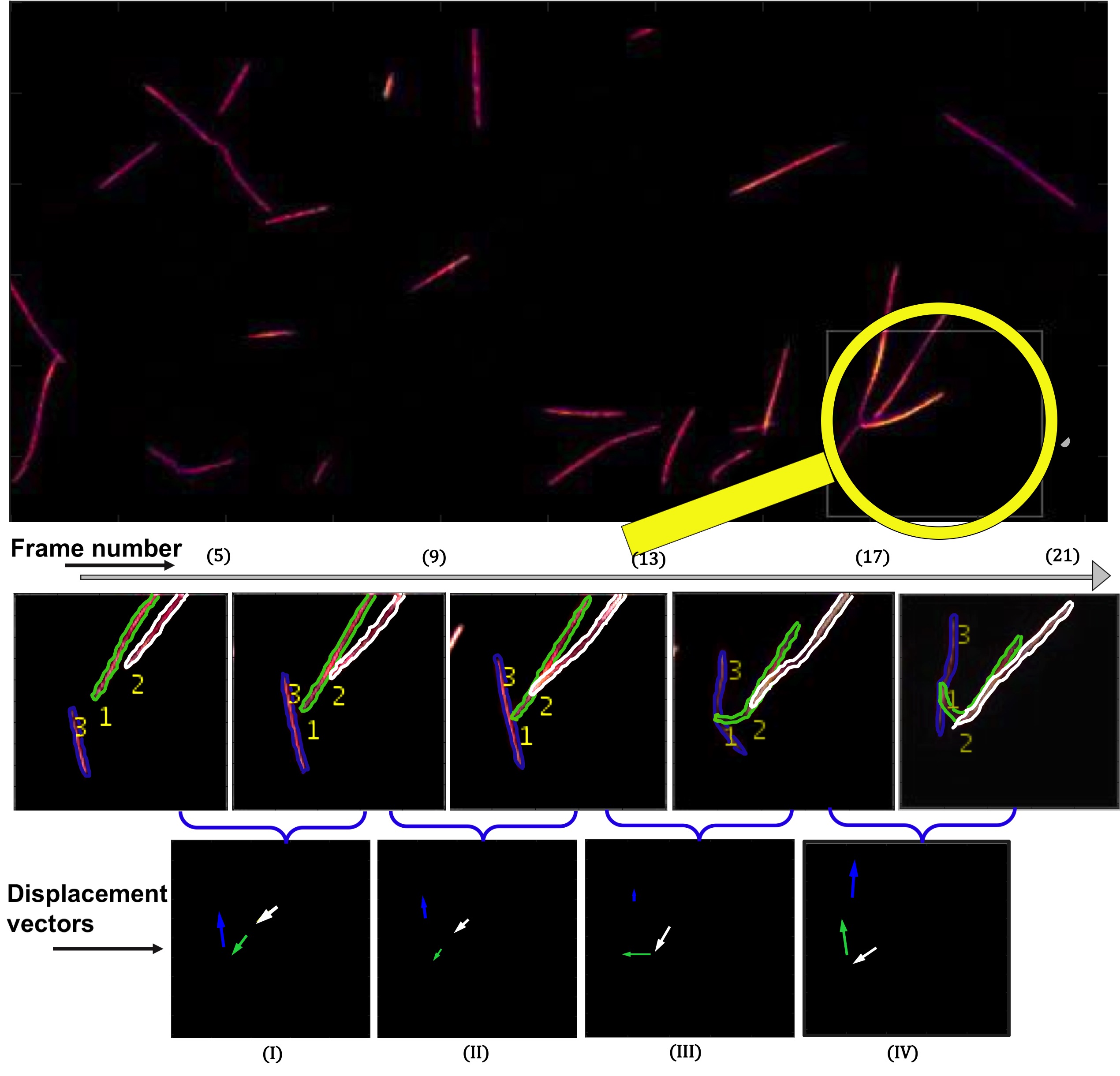}
\caption{\textcolor{black}{A single $256\times512$ pixel image of gliding microtubules (first row). MTs are pseudo-colored to reflect differences in intensity, with warmer colors indicating higher intensity values and cooler colors indicating lower intensity values. The middle row depicts a zoomed-in, time evolution of the area highlighted within the magnifying glass (sampling interval per frame was 250 ms). The bottom row shows the frame-by-frame displacement vectors} of MTs 1, 2, and 3 in green, white, and blue respectively. A thorough version of these frames are included in supplementary materials section.} 
\label{Sample Images}
\end{figure*}
\textcolor{black}{Qualitative analysis of our image sequences indicates that sudden changes in MT velocity, in terms of direction \textit{or} amplitude, often occur concurrently with obvious MT-MT collision and interaction. Such an event is depicted in Figure \ref{Sample Images}, where interaction among three MTs results in obvious change their velocities. MT velocity is defined as a motion vector with respect to the leading end of the MT (i.e. its \textit{head})} disregarding its dynamic instability~\cite{Smasoudi2}. \textcolor{black}{As can be seen in Figure~\ref{Sample Images}, velocity changes are manifested as changes} of either amplitude (MT3, blue vector in II and III), direction (MT1, green vector in III and IV), or both (MT1, green vector in II and III).

\textcolor{black}{To characterize these changes}, one must track individual MTs in sequential frames. However, instance-level MT tracking problem is challenging for the following reasons: 
\begin{itemize}
\item low diversity in MT appearance,
\item time-varying nature of the features as a result of dynamic instability and photobleaching,
\item abrupt appearance/disappearance of MTs (caused in part by the use of TIRF microscopy which illuminates only the \textcolor{black}{first 100nm of depth from the coverslip surface}), and
\item unexpected changes in MT \textcolor{black}{shape}, ascribed to MT interaction and collision.
\end{itemize}

To address these challenges \textcolor{black}{and limitations inherent to the existing methods}, we propose a generic solution composed of two distinct but complementary parts. \textcolor{black}{The first part of our solution introduces a novel instance-level MT segmentation method (at each frame). The second part tracks MTs along time-lapse images by utilizing these instance segmentation results in a data-association framework. This research is a big step toward development of an analysis platform tool which enables biologists to characterize the effect of MT-MT interactions on MTs velocity.}
\subsection{Related works}
Early visualization of MTs started with time-lapse images captured from \textcolor{black}{either cells injected with fluorescently-labeled tubulin or those expressing fluorescent protein-tubulin fusions~\cite{goodson2004live,semenova2007fluorescence}.}
 \textcolor{black}{Application of this method was typically restricted to the periphery of interphase cells where MT density was sufficiently low to capacitate the high contrast imaging of individual MTs. The \textit{in vivo} exploration of MTs, evolved considerably with the use of fluorescently labeled +TIPs, proteins that bind specifically to the growing MT plus ends~\cite{tirnauer2000eb1}}. Tracking the +TIPs revealed descriptive parameters of dynamic instability like MT nucleation rate and growth speed~\cite{gatlin2009spindle, smal2008particle, matov2010analysis}. The computational analyses for +TIPs tracking, are principally derived from multiple particle tracking algorithms in contrast with few solutions based on dense field motion detection~\cite{fortun2015optical}. 
 
 Literature on multiple particle tracking implies two steps of (1) recognition of the relevant particles, and (2) associating the segmentation results~\cite{applegate2011plustiptracker, smal2008particle, matov2010analysis}. The performance of each step directly affects the quality of the obtained spatiotemporal trajectories.
 \begin{figure*}[t]
\centering
\includegraphics[width=0.85\textwidth]{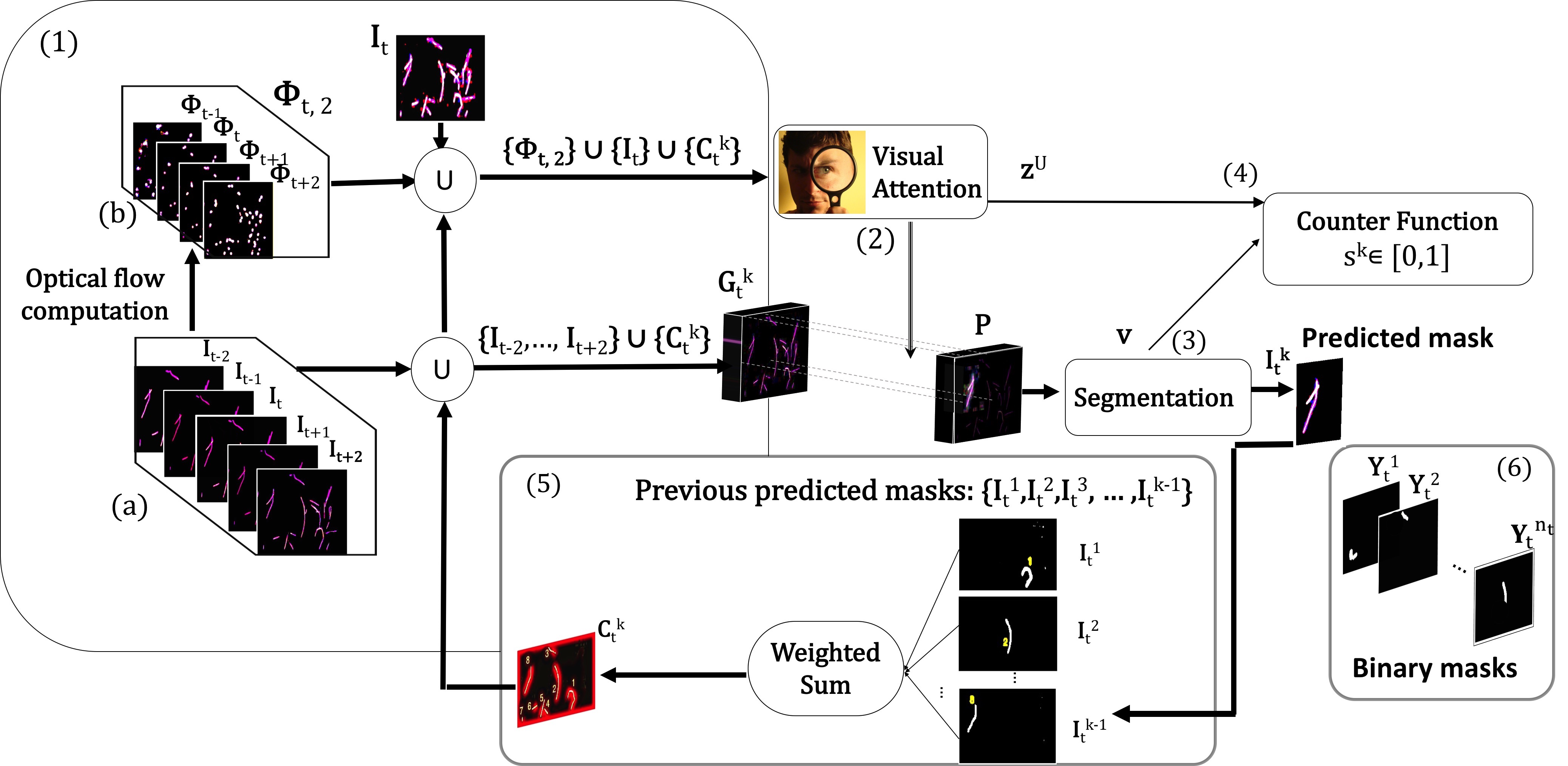}
\caption{Overview of the proposed method to segment instance $k$ MT at time $t$. In (1) the current frame, $\textbf{I}_{t}$, and 2L (L=2) number of its surrounding frames along with $\textbf{C}^{k}_{t}$ (${k}-1$ already segmented instances at the current frame), form (a) the input group $\textbf{G}^{k}_{t}$, (b) 2L+1 frames reform to a set of 2L (L=2) OFs. These OFs, current frame, and $\textbf{C}^{k}_{t}$ configure a group to supply the visual attention, (2) visual attention determines where the network should look for the $k^{th}$ instance, (3) segmentation recognizes $k^{th}$ instance inside the attention region, (4) evaluates the performance of visual attention and segmentation elements to return a score for segmentation quality, if the score drops under a certain threshold, network stops iterating over the same frame, resets itself and moves forward to the next frame, $\textbf{I}_{t+1}$, (5) collects the already segmented instances ${1:k-1}$, linearly combine them and feeds the result back to the input, (6) represents the layers of binary masks for individual MT labels.} 
\label{General}
\end{figure*}

 \textcolor{black}{Literature on segmentation methods from pre-deep learning era is vast: clustering,  region growing, morphological filtering, template matching, wavelet decomposition, graph and fuzzy set algorithms~\cite{zaitoun2015survey}. However, only a few of these methods are applied to the context of sub-cellular particle detection as well as MT segmentation (see a comprehensive review~\cite{kervrann2016guided}}. Among such methods, the oldest one is thresholding that takes advantage of \textcolor{black}{differences in fluorescent intensity between the objects being tracked and the background. Previously in \cite{Smasoudi2}, we employed a global threshold value via Otsu's method to segment MTs. Debated by~\cite{applegate2011plustiptracker}, global thresholding alone cannot afford the ideal segmentation in microscopy images where noisy background, poor image quality, and heterogeneous particles exist. Various pre-processing ideas have been developed to partially solve such difficulties. For instance, authors in~\cite{matov2010analysis} applied a Gaussian band-pass (BP) filter before global thresholding. Similarly, \cite{smal2010quantitative} used Gaussian denoising and morphological operation followed by thresholding for +TIPs segmentation. To avoid the shortcomings of global thresholding, \cite{applegate2011plustiptracker} applied local thresholding where local thresholds are the local maximum of the BP filtered image. Thresholding was usually used to generate seed points, feeding an additional algorithm for more precise delineation. Such algorithms in literature include either region growing~\cite{matov2010analysis} or watershed segmentation~\cite{applegate2011plustiptracker}. Despite these engineering efforts, over and under-segmentation problems persisted. In another attempt, \cite{applegate2011plustiptracker} employed a post step of template matching similar to~\cite{matov2010analysis} to  benefit from the shape of desired objects in cell. In a different line of research, ~\cite{zhang2007multiscale,olivo2002extraction} used wavelet decomposition for object detection. Regardless of the specifics of the approach used, none of these methods allows to segment MTs in a sequence of time-lapse images for ultimate purpose of tracking. This inability is due to the time-varying nature of the image intensities, caused by photobleaching, molecular-level processes, or unique sub-cellular dynamics.} 
 
Deep learning based instance-level segmentation has became a rapidly growing area of study in recent years. Popular methods such as~\cite{Hayder2017,Li2017,Bai2017,Romera-Paredes2016,Dai2016} mostly perform simultaneous instance-level and semantic segmentation for both classification and segmentation. Conventionally, the regulation of relevant strategies is composed of a box proposal followed by parallel processing for classification and detailed segmentation. The rationale for using the bounding box is due to the strong coupling between segmentation and object detection: once the object is found, delineation can be performed within each box (detected object). Mask R-CNN~\cite{he2017mask}, and its extensions,~\cite{Liu2018}, and~\cite{Hu2018} are among the recent works with great potentials in this stream. However, training this type of algorithms demands huge collection of labor-intensive annotations which is major drawback in case of biomedical applications.

Several data association approaches exist in literature. These algorithms optimize the association cost among the obtained results from two~\cite{ jaqaman2008robust, godinez2015tracking}, or more frames (multiple succeeding frames~\cite{ chenouard2013multiple, jaiswal2015tracking} or larger batches of frames with more complex graph pruning techniques~\cite{ racine2006multiple,roudot2017piecewise}). Several challenges emerge in assigning the segmented objects from individual frames to each other. Among these, the problem of low signal to noise ratio (SNR) was resolved by the application of probabilistic approaches~\cite{smal2008particle}. Even in presence of adequate SNR, attributing the suddenly appearing/disappearing particles to their true trajectories was a real struggle. \textcolor{black}{+TIP imaging exemplifies this issue where its inability to visualize MTs during the pause and shrinkage phases}, necessitates extra processing \textcolor{black}{~\cite{matov2010analysis,Applegate_thsis,applegate2011plustiptracker,jaqaman2008robust}}. To compensate for MTs missing phases computationally, an algorithm was proposed by~\cite{jaqaman2008robust}. The \textit{plusTipTracker} software package~\cite{applegate2011plustiptracker} was designed based on this algorithm to trace MT plus ends in +TIP images. The heterogeneous growing patterns exhibited by +TIPs was yet another challenging aspect. Interacting multiple model filtering, piecewise-stationary motion modeling, and piecewise-stationary multiple motion Kalman smoother are the latest studies that incorporate the Bayesian prediction power to optimize assignment~\cite{racine2006multiple,smal2010MT,roudot2017piecewise, Smasoudi2}. There is a complete literature review on the most common data association techniques in particle tracking applications in~\cite{smal2015quantitative}. It is known that false negative rates are far more problematic to data association than imperfect detection. By avoiding mis-detections (reducing the false negatives), there is no need to use sophisticated multi-frame linking techniques~\cite{smal2015quantitative}. The most recent development in this area is the application of deep learning  to the problem of data association in multiple particle tracking~\cite{yao2018deep}.


\textcolor{black}{Sub-cellular particle tracking can be be potentially addressed by dense motion detection. Optical flow (OF) is a basic features to describe motion in a dense field~\cite{fortun2015optical}. There are several strategies for OF computation, some of which were applied to microscopy images.~\cite{liu2014optical} and~\cite{delpiano2012performance} used OF for cell tracking and motion estimation of cellular structures. In this regard, Horn and Schunck OF (HS-OF) computation method is a global approach based on two assumptions: \textit{gray value constancy} and \textit{smooth flow of the intensity values}. Later, \cite{delpiano2012performance} utilized additional constraints to extend HS-OF to \textit{combined local global} method. Tracking solutions that incorporate OF computation \textit{solely} have many drawbacks: losing small moving structures due to the coarse to fine decomposition, over-smoothing motion discontinuities caused by variational optimization, and having difficulty in dealing with illumination changes~\cite{fortun2013aggregation}. To address these limitations, a patch-based two-step aggregation framework was proposed to estimate the motion patterns of cellular structures~\cite{fortun2013aggregation}.} 

\textbf{Research gap}:
\textcolor{black}{Literature on MTs is particularly focused on descriptive parameters of dynamic instability. Our project presents a new problem toward estimating the (translational) velocity of MTs during \textit{in-vitro} gliding assays. We fulfill this task through a deep instance-level MT segmentation and associated tracking method. Individual MT velocity estimation demands identification of each single MT which is performed using instance-based segmentation. Unlike other instance-level segmentation methods in computer vision applications, we focus on MTs only (foreground) to avoid category dependant computations, extensive number of parameters, and numerous costly annotations. We take advantage of \textit{attention} modeling to improve segmentation results and allow separation of MTs. In contrast to~\cite{fu2017look,he2017fine,peng2018object} who used attention to get fine-grained details of a single instance, our study uses attention to exploit the spatial relation among different MT instances in an image. After identifying the attention region(s), the exact instance mask is thoroughly segmented.} 

 \section{Methods}
\subsection{Overview of the proposed method}
Our method can be best described under two headings: Part 1) instance-level MT segmentation at each frame, Part 2) MT association among successive frames. The first involves segmentation which becomes extremely challenging when MTs \textit{overlap} or \textit{collide}. To alleviate this, we present a new instance-level segmentation algorithm utilizing a recurrent neural network (RNN).  The segmentation procedure is guided by a novel visual attention module repeatedly processing a single frame to segment its MT contents. This module facilitates efficient delineation of individual MTs even when MT-MT interactions exist. We describe our solution in five steps.  Step 1 is the data preparation module at time $t$: the current frame, a sequence of its neighboring frames or their respective OFs, and weighted sum of already segmented instances at the current frame are grouped together as input. Step 2 describes the visual attention module which proposes \textit{where to focus} for segmentation. Step 3 includes the segmentation unit for each instance inside the area  suggested by Step 2. Step 4 is a counter function validating Steps 2 and 3 to decide when to stop iterating over the same frame. 
In Step 5, the most recent segmented instance joins all the previously recognized instances from the same frame and the weighted sum is fed back to the input. The algorithm repeats the same procedure on the same frame, until the counter in Step 4 signals to stop. At this level, we obtain instance-level segmentation results at the $t^{th}$ frame. So MT instances can be segmented in every single frame following this procedure. Later in Part 2, \textcolor{black}{we use Hungarian algorithm~\cite{kuhn1955hungarian}, to assign the segmented MT instances along  every pair of the succeeding frames.} As a consequence, we get trajectories of MTs along the frames that promote MT velocity estimation. Figure~\ref{General} shows the flowchart of our segmentation platform. 

To the best of our knowledge, this is the first study exploring the problem of instance-level MT velocity estimation with a deep learning algorithm. Due to the limited and extremely hetergeneous nature of our real data, we first create a simulated data based on statistics derived from the actual time-lapse microscopy images of MTs. Such simulated data provides a means to pre-train our deep learning framework and optimize its hyperparameters before fine-tuning on our limited real data. 


\subsection{Problem statement}
 The problem throughout this paper is to estimate the translational velocity of each individual MT along the subsequent frames in a given set $\textit{\textbf{I}}=\{\textbf{I}_{1}, \textbf{I}_{2}, ..., \textbf{I}_{T}\}$. While all these frames share similar dimensions: $\text{H}_1\times\text{W}_1\times 3$ \textcolor{black}{(3 RGB channels)}, each may contain a different number of instances due to MTs sudden appearance/disappearance, marginal entrance and egress. The true number of MT instances at the ${t}^th$ frame is ${n}_{t}$ which are denoted by binary ground truth masks: $\{\textbf{Y}^{1}_{t}$, ..., $\textbf{Y}^{n}_{t}\}$.
 
As previously explained, our method is composed of two parts: for Part 1, we propose a configuration that sequentially goes through single frames from $\textit{\textbf{I}}$ to perform instance-level MT segmentation at each frame. As a result, we obtain ${m}_{t}$ binary masks of MT instances at the ${t}^th$ frame that are represented by: \{$\textbf{I}^{1}_{t}$,...,$\textbf{I}^{m}_{t}\}$. These obtained masks are compared against ${n}_{t}$ binary ground truth masks to evaluate our segmentation performance at this frame. Once we segment all MTs in each frame along a sequence of successive frames, we move on to Part 2. For this Part, we associate the results from each pair of successive frames $\textbf{I}_{t}$ and $\textbf{I}_{t+1}$ to recognize an individual path for each MT and estimate its velocity.  
We want our network to learn to segment instances with conflicting areas. To facilitate this, we append all binary ground truth instances through their third dimension and form a 3-D label tensor $\textbf{Y}_{t}$. Using a 3-D label while training, enables our network to account for the overlapped area among individual instances at $\textbf{I}_t$ .

 \subsection{Part 1: Instance-level MT segmentation in a single frame}
\begin{figure}[!h]
\centering
\includegraphics[width=.45\textwidth]{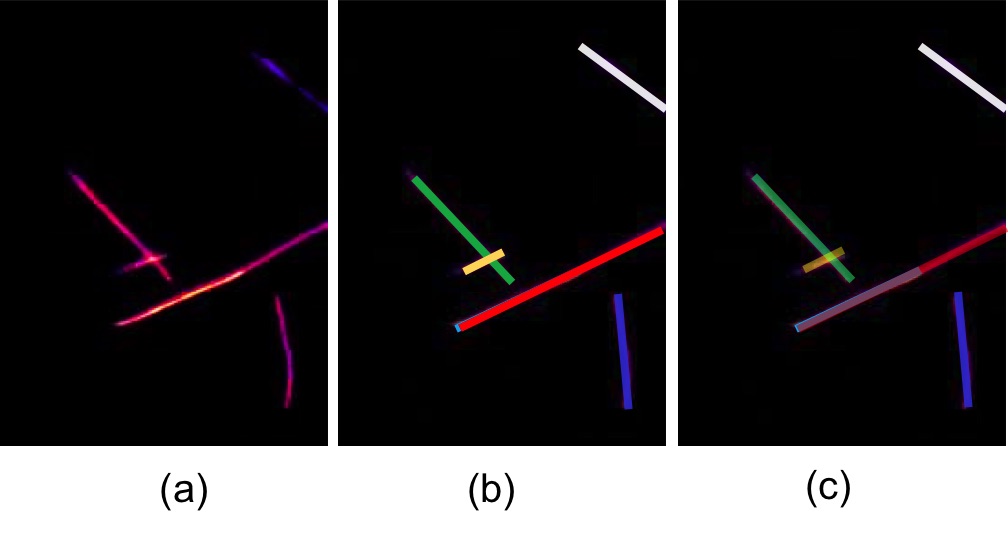}
\caption{\textcolor{black}{(a) original scene, (b) conventional segmentation (no overlap among the computed instances), (c) our desired segmentation where computed segments may considerably overlap (transparent colors are used to represent the overlapping among multiple instances).}}
\label{segmentation}
\end{figure}

 Inspired by ~\cite{ren2017end} and~\cite{hu2017maskrnn}, we present a new system of attention to accurately segment individual MTs. The attention module generates particular Gaussian kernels to specify \textit{where to look} for the next instance. These kernels blur out the exterior and enhance the interior of the attention area. Later, the segmentation network extracts MT(s) from the suggested region. \textcolor{black}{Unlike~\cite{ren2017end}, segmentation herein is our intermediate goal to realize instance-level MT velocity estimation along time-lapse images at the end. Additionally, MTs overlap considerably hence there is a need for specific type of instance segmentation as demonstrated by Figure~\ref{segmentation}.}

\begin{figure}[!h]
\centering
\includegraphics[width=.48\textwidth]{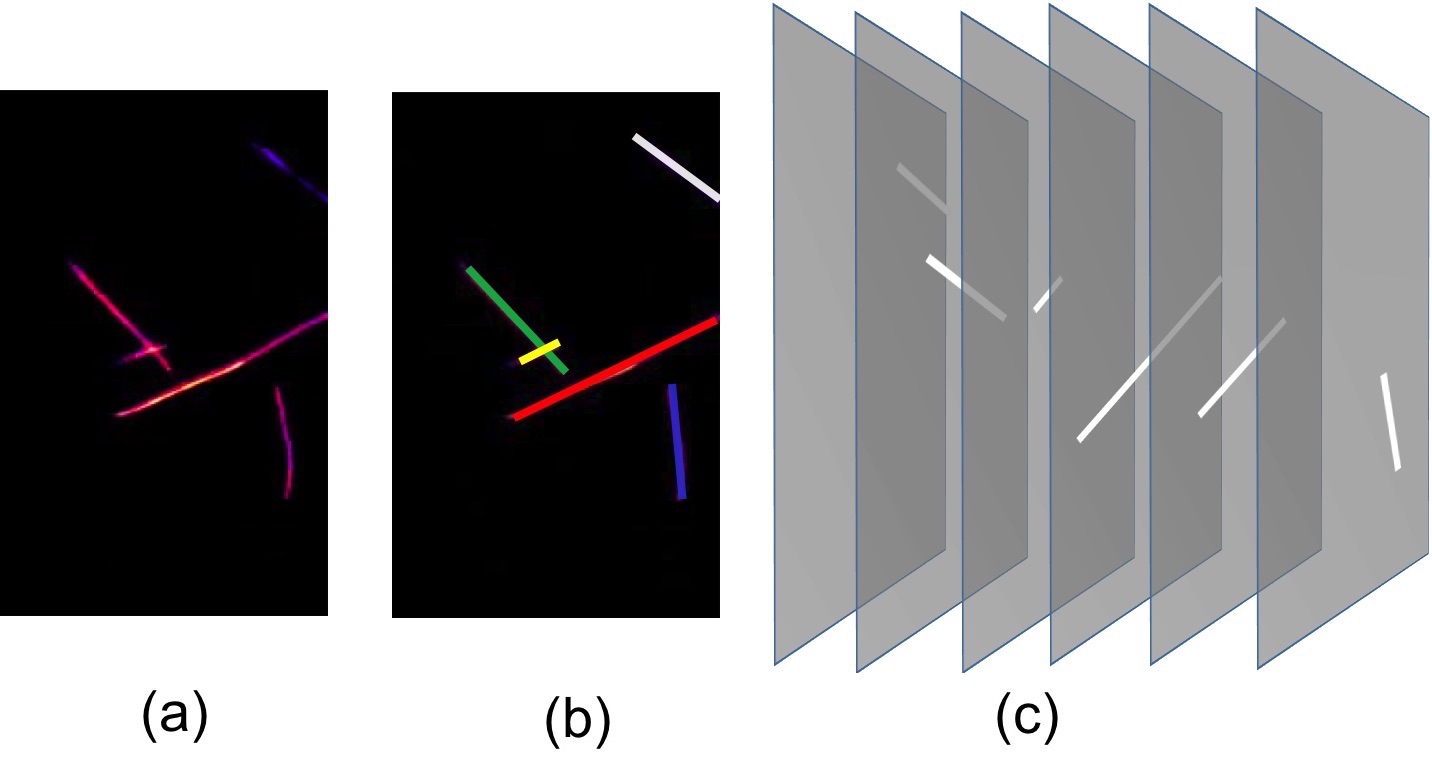}
\caption{\textcolor{black}{(a) original scene, (b) labeling according to~\cite{ren2017end} where a distinct color is assigned to the upper most instance at each location, (c) our desired labeling strategy which enables the network to learn about the whole instance regardless of other objects overlapping.}}
\label{labels}
\end{figure}

 Our work has improved~\cite{ren2017end} in three major ways. We use 3-D labeling to secure a comprehensive segmentation in case of overlapped instances: layers of ground truths  $\textbf{Y}_{t}^{k}$ for all individual instances with potential common areas are appended through their third dimension and form a 3-D label tensor $\textbf{Y}_{t}$ as is depicted by Figure~\ref{labels}. 
 
 \begin{figure*}[!h]
 \centering
 \includegraphics[width=0.9\textwidth]{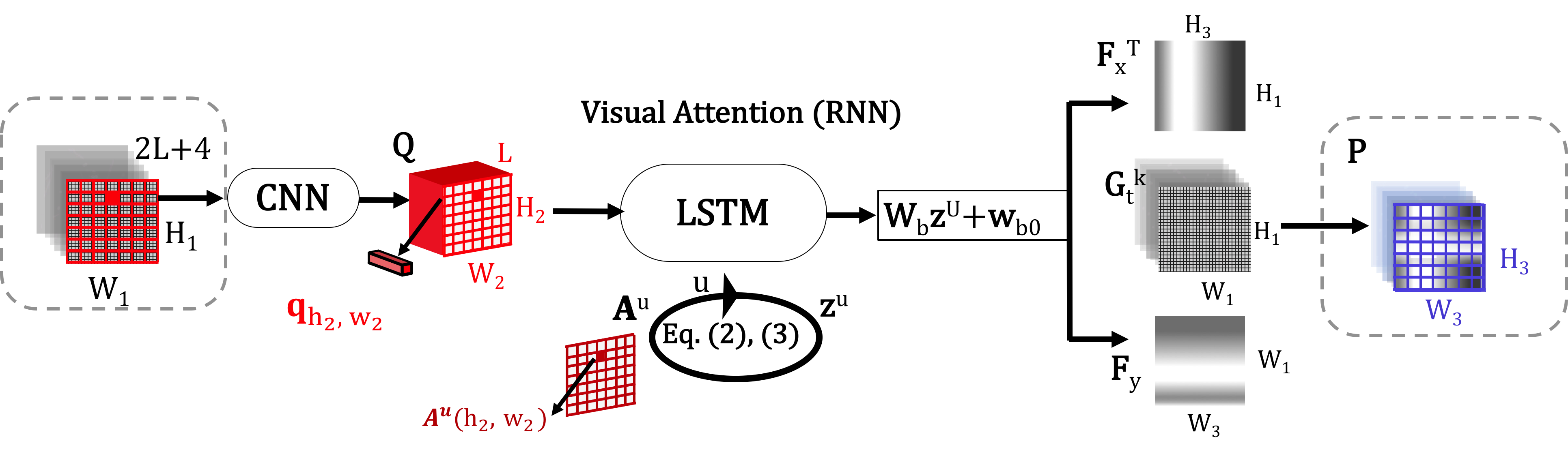}
  \caption{Visual attention module: CNN produces a $D$-dimension feature vector $\textbf{q}_{{h}_2,{w}_2}$ for each tile, LSTM iterates $\text{U}$ times and assigns a contribution level $\mathbf{A}^{{u}}({h}_2,{w}_2)$ to the respective tile at its ${u}^{th}$ repetition, after convergence of the LSTM, its recent hidden state implies two vertical and horizontal Gaussian kernels. These kernels together propose the attention region.} 
  \label{Tile}
\end{figure*}

In addition, we extend~\cite{ren2017end} into the temporal domain to collect sufficient cues for segmenting the concealed areas. We utilize former and future frames to obtain location- and appearance-related evidences that support our algorithm to segment overlapping instances. Finally, we improve the long short-term memory (LSTM) implementation from a fixed number of iterations in~\cite{ren2017end} to a conditional convergence. Using a constant number of iterations can critically restrict the LSTM performance in proposing a new attention region.

 To elaborate our proposed strategy for instance-level MT segmentation at frame $\textbf{I}_{t}$, we assume our network to begin its ${k}^{th}$ iteration in an attempt to find the ${k}^{th}$'s instance. Terms ${\textbf{G}_{t}}^{k}$ (defined in Equation \ref{Eq0}) and ${\textbf{I}_{t}}^{k}$, respectively describe the input and output of our algorithm at time $t$ for instance $k$. With these assumptions, the overall structure of our network has the following elements:

\subsubsection{Input}
\textcolor{black}{To provide temporal information at the input, we use the frames in neighborhood of the current frame. We also include ${\textbf{C}_{t}}^{k}$, a weighted average of all previously segmented instances at the current frame (see Equation \ref{eq101} for details), to accommodate reasoning about a new instance. A sequence that contains L number of (former) frames, current frame, and L number of (future) frames alongside the most updated ${\textbf{C}_{t}}^{k}$, form a tensor to supply the input group, $\textbf{G}^{k}_{t}$:}
\begin{equation}
    \textbf{G}^{k}_{t}=\{\textbf{I}_{t-L},...,\textbf{I}_{t},..., \textbf{I}_{t+L},{\textbf{C}_{t}}^{k}\}.
\label{Eq0}
\end{equation}
\textcolor{black}{Tensor $\textbf{G}^{k}_{t}$ is the input to both the visual attention and the segmentation block. Yet, we examine another version of data preparation at the input, where we substitute the neighboring frames with their respective OFs to feed the visual attention. Using OF provides visual attention with indicative features of the motion vectors. For this purpose, we follow the work of Liu~\cite{liu2009beyond} to compute OF from each pair of successive frames within the neighborhood of 2L+1. The resulting set of OFs, $\mathbold{\Phi}_{t,\textit{L}}$, the current frame, and ${\textbf{C}_{t}}^{k}$ all together set up an input for visual attention. Experimental results for this trade-off are explained later in Section III.}

\textcolor{black}{We will henceforth drop the subscript $t$ and superscript $k$ for clarity. All subsequent terms that describe quantities inside the visual attention, segmentation, and counter function are assumed to at time $t$ for instance $k$.}

\subsubsection{Visual attention}
\textcolor{black}{Our proposed attention network contains a convolutional neural network (CNN) followed by LSTM in the \textit{spatial-domain} as depicted by Figure~\ref{Tile}}. The goal is to provide the region of interest for a constrained segmentation. \textcolor{black}{First, the CNN passes the input volume through successive layers of convolutional filters and max pooling such that the $\text{H}_1\times\text{W}_1$ area of the input group is reduced to $\text{H}_2\times\text{W}_2$ non-overlapping tiles. Hence, the CNN generates a feature tensor $\textbf{Q}$ of shape $\text{H}_2\times\text{W}_2\times\text{D}$, where each spatial location is a ${D}$-dimensional feature vector expressed by $\textbf{q}_{{h}_2,{w}_2}$, as illustrated in Figure~\ref{Tile}.} 

\textcolor{black}{Next, we utilize LSTM to model the spatial causality among multiple instances in a frame; i.e., it uses the spatial features of former instances: $\{1,...,k-1\}$ segmented at the current frame to estimate the area of attention for ${k}^{th}$ instance at the same (current) frame.}
Upon receipt of the feature tensor, the LSTM begins iterating to find those tiles in $\textbf{Q}$ that contribute the most to the attention region. After any given ${u}^{th}$ iteration, LSTM produces a hidden state vector $\textbf{z}^{{u}}$ and a 2-D matrix, $\mathbf{A}^{{u}}$ of size $\text{H}_2\times\text{W}_2$. Every entry in matrix $\mathbf{A}^{{u}}$ expresses the level of contribution to the attention region for its respective tile in $\textbf{Q}$. Initiating with equal involvement of all tiles, they gradually fine-tune (Eqs.~\ref{eq2} and~\ref{eq3}):
\begin{align}
\textbf{A}^{{u}}=\begin{cases}
           1/(\text{H}_2\times \text{W}_2), & \text{if} \quad {u}=0,\\
           \text{MLP}(\textbf{z}^{{u}}), & \text{otherwise,}
		\end{cases}
\label{eq2} 
\end{align}

\noindent where MLP(.) in this equation denotes a single hidden layer multi-layer perception with 5 hidden units, and 
\begin{align}
\textbf{z}^{{u}}=\begin{cases}
               0,   & \text{if}   \quad {u}=0,\\
              \text{LSTM}\bigg(\textbf{z}^{{u-1}}, \sum\limits_{{h}_2,{w}_2}{\mathbf{A}^{{u-1}}({h}_2,{w}_2)\textbf{q}_{{h}_2,{w}_2}}\bigg), & \text{otherwise.}
            \end{cases}     
\label{eq3}
\end{align}
 The LSTM repeats until each element in $\mathbf{A}^{{u}}$ converges \textcolor{black}{or $u$ reaches a set maximum number}. We refer to the last iteration number as $\text{U}$ and use it as an upper index to specify the ultimate hidden state $\textbf{z}^{\text{U}}$. Using a linear transformation of vector $\textbf{z}^{\text{U}}$, we compute description parameters of the attention region as shown in Eq.~\ref{eq4}. These parameters define mean and standard deviation of two Gaussian kernels $\textbf{F}_x$ and $\textbf{F}_y$ along the $x$ and $y$ axes:
\begin{align}
[\mu_{x},\mu_{y},\sigma_{x},\sigma_{y}]^\intercal=\textbf{W}_{b}\textbf{z}^{\text{U}}+\mathbf{w}_{b0}.
\label{eq4}
\end{align}
Gaussian kernels ($\textbf{F}_x$ and $\textbf{F}_y$)are calculated using,
\begin{align}
& \textbf{F}_x({h}_1,{h}_3)=\frac{1}{\sigma_{x}\sqrt{2\pi}}\exp{-\frac{({h}_1-\mu_{x})^2}{2{\sigma_{x}}^2}}\label{eq5},\\
&\textbf{F}_y({w}_1,{w}_3)=\frac{1}{\sigma_{y}\sqrt{2\pi}}\exp{-\frac{({w}_1-\mu_{y})^2}{2{\sigma_{y}}^2}}.
\label{eq76}
\end{align}
Then, we transform the input $\textbf{G}^{k}_{t}$ into $\textbf{P}$ (Eq.~\ref{eq7}).
\begin{align}
\textbf{P}={\textbf{F}_x}^\top \textbf{G}^{k}_{t}\textbf{F}_y.
\label{eq7}
\end{align}
Doing so, we fulfill two tasks simultaneously: first, intensifying the attention area, while attenuating the rest of the frame. Second, re-sampling the $\text{H}_1\times \text{W}_1$ area of input group into ${\text{H}_3}\times {\text{W}_3}$ for magnified details in $\textbf{P}$ ($\text{H}_3<\text{H}_1$ and $\text{W}_3<\text{W}_1$). In other words, pixels from the original current frame contribute to each pixel in the attention region according to matrices $\textbf{F}_x$ and $\textbf{F}_y$.

\subsubsection{Segmentation}
To segment an object in the attention region, we apply a back-to-back Encoder-Decoder similar to~\cite{noh2015learning}. This design transforms our attention-magnified input group $\textbf{P}$ into a $D'-$dimensional feature vector $\textbf{v}$ first. Later, $\textbf{v}$ is decoded into a pixel-wise prediction map $\hat{\textbf{P}}$. To have the segmentation result in a comparable size with the original image, we undo the effect of the Gaussian kernels: 
\begin{align}
\textbf{v}&=\text{Encoder}\big(\textbf{P}\big),\nonumber  \\ \hat{\textbf{P}}&=\text{Decoder}\big(\textbf{v}\big),\label{eq81}\\
\textbf{I}^{k}_{t}&={\textbf{F}_x}\hat{\textbf{P}}{\textbf{F}_y}^\top,\nonumber 
\end{align}
where again all of these quantities are being computed at given time $t$ and instance $k$.
\subsubsection{Counter function}
\begin{figure}[!ht]
\centering
\includegraphics[width=0.4\textwidth]{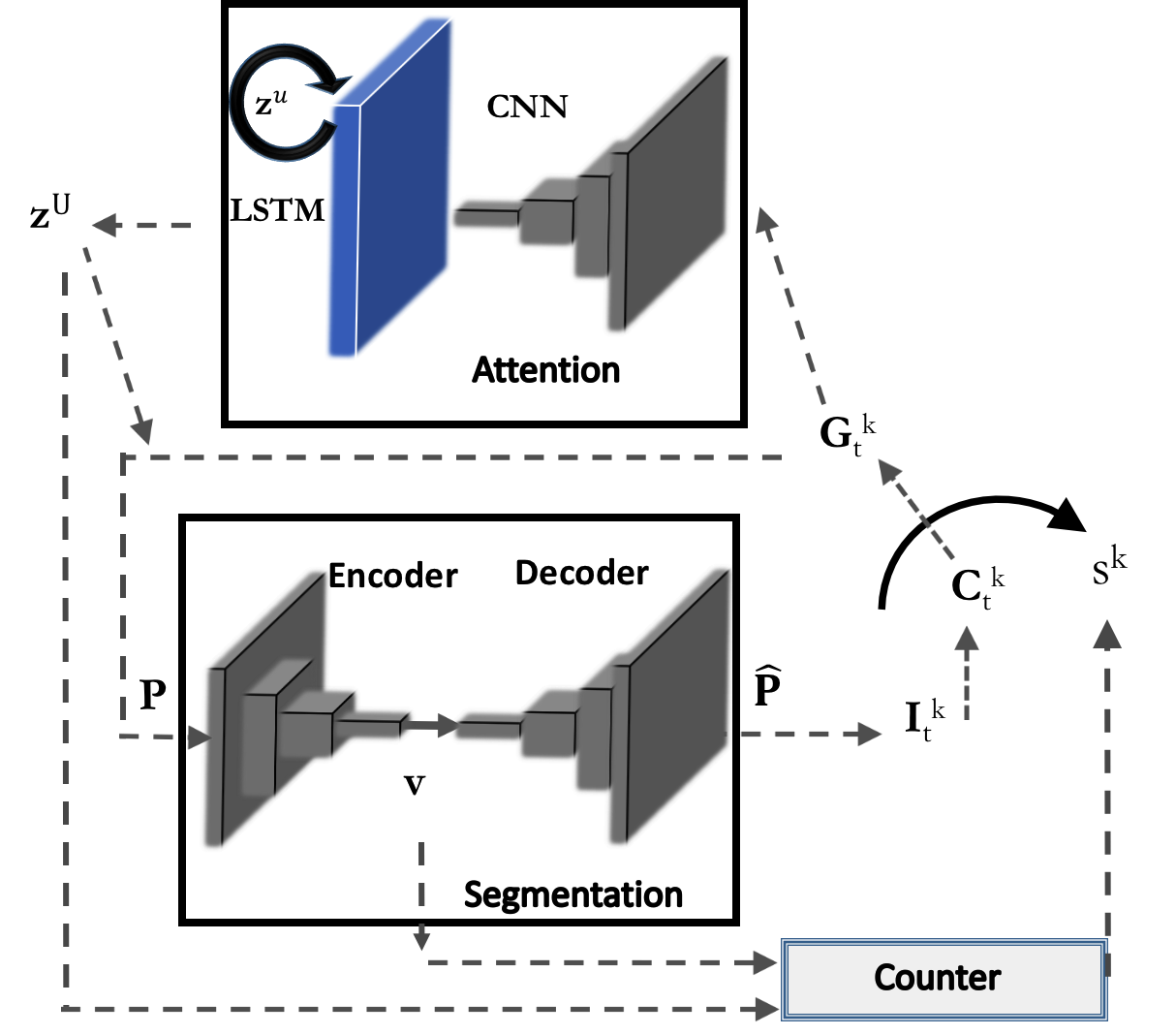}
\caption{Structural details of the deep network layers; \textcolor{black}{The CNN in this design has 8 layers, each layer is made of a Conv layer ($3\times3$), a max pooling layer followed by a RELU function. Filter depths and max pooling filter sizes in these layers are {8,8,16,16,32,32,64,64} and {1,1,2,2,1,2,2,2} respectively. Encoder, decoder have a forward and backward paths respectively through a network of 6 layers, each of which with a Conv layer ($3\times3$), a max pooling layer followed by a RELU function, filter depths and max pooling filter sizes in these layers are {8,8,16,16,32,32} and {1,2,1,2,1,2} respectively}. Counter resets the network to move froward to the next frame.} 
\label{Struct}
\end{figure}

A critical addition to this architecture is the counter function which determines whether the attention region includes an instance or not. The counter function is made of a fully connected layer and a Sigmoid function. This module takes in a concatenation of two vectors $\textbf{z}^{\text{U}}$ (the most updated RNN hidden state) and $\textbf{v}$ (encoder output) to generate a score value $s^k$ (see Figure~\ref{Struct}).
We train the weights in the counter function so that the value of the $s^k$ lies in the interval of $[0,1]$. A higher score expresses more certainty toward the instance segmentation. As is depicted by Figure~\ref{Struct}, the counter function acts like a switch during the test. If $s^k$ surpasses a certain threshold, the network counts the segmented instance as a successful attempt, moves on to segmenting the next instance $k+1$ for the same frame $\textbf{I}_{t}$. Otherwise, an unqualified instance segmentation forces the network to stop iterating through the same frame, reset itself and step forward to the next frame, $\textbf{I}_{t+1}$ along the time-lapse images. 
\subsubsection{Feedback} 
Each segmented instance joins all previously segmented instances to constitute ${\textbf{C}_{t}}^{k}$. 
Feeding this weighted average as part of the input set (Eq.~\ref{eq101}) into our network facilitates the segmentation of future instances in two ways: reduces the chance of selecting a region among already assigned areas, and provides a prior to the network based on the potential relation between various instances:
\begin{align}
{\textbf{C}_{t}}^{k}=
            \begin{cases}
              \textbf{0},   & \text{if}   \quad {k}=1,\\
              \frac{1}{{k}-1}\sum\limits_{{j}=1}^{{j}={k}-1}{\textbf{I}^{j}_{t}} & \text{if} \quad {k}>1.
            \end{cases}     
\label{eq101}
\end{align}
 
\begin{figure}
\centering
\includegraphics[height=.3\textwidth]{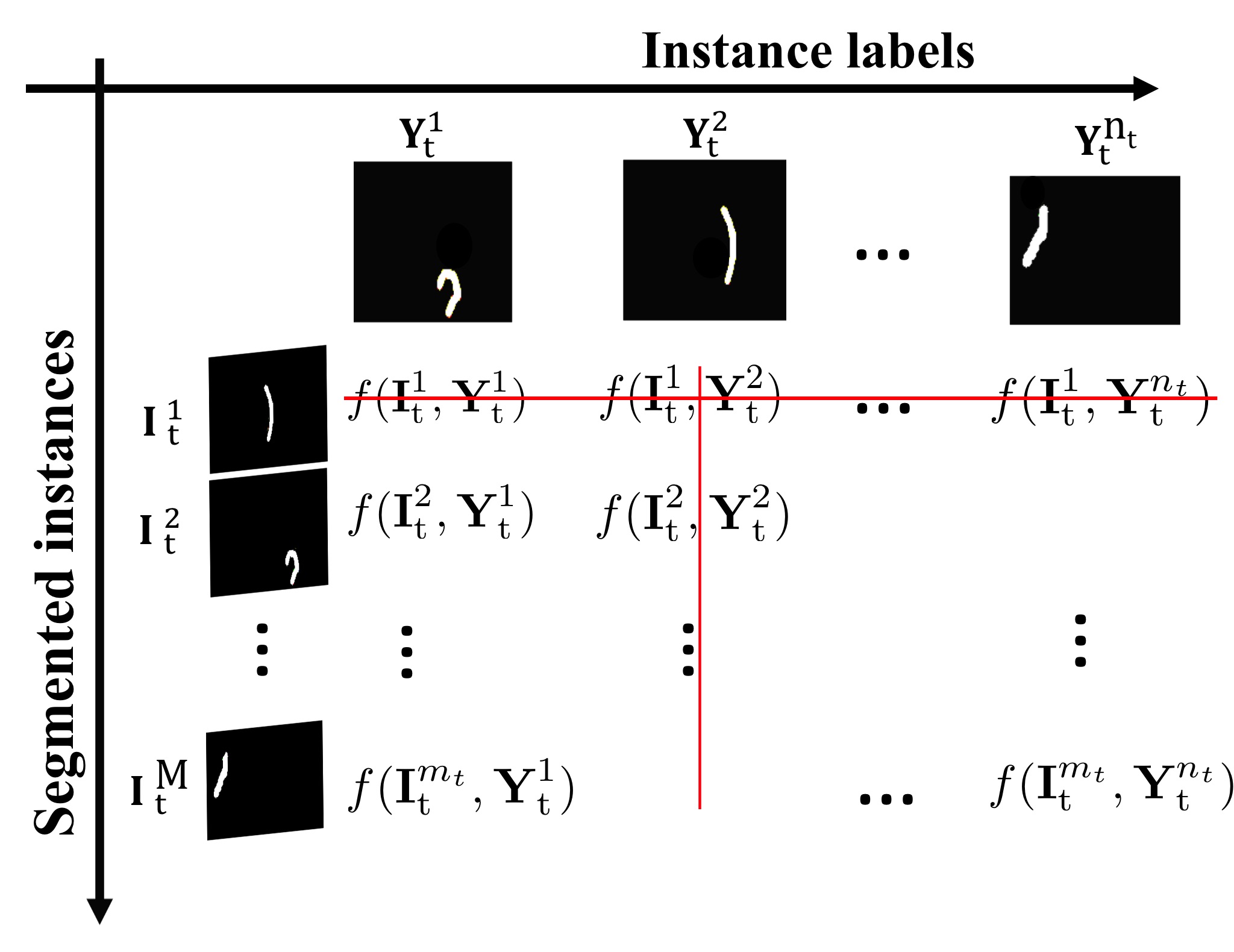}
\caption{The cost matrix of the Hungarian algorithm which measures the similarity between the segmented instances and instance labels (the higher values express more similarity).} 
\label{Hung_alg}
\end{figure}

\textbf{{Training for segmentation}}: Our proposed instance-level segmentation algorithm closely ties the segmentation and attention networks to each other. However, the level of dependency varies among these two networks. The segmentation performance is directly determined by the attention accuracy but the segmentation results only provide extra guidance to the attention network. Such coupling forces us to 
implement the training procedure in two stages. First, we ignore the segmentation network and train the attention network only. Second, we train the whole network by fine-tuning the attention weights and optimizing the segmentation network from scratch. At this stage, feeding back the premature segmentation results into the network can be misleading. Therefore, we define a "tuning-knob" parameter. This parameter enables us to \textcolor{black}{feed the ground truth instance into the network and gradually replace it with the results from the segmentation network as the training progresses}. Since the counter function must be trained to distinguish successful performance, in both training stages, we force our algorithm to iterate $M$ times through each frame, where $M$ is determined from Eq.~\ref{eq100}:
\begin{equation}
M =\max\limits_{1\leq t\leq T}\big({n}_{t}\big),
\label{eq100}
\end{equation}
where again ${n}_{t}$ is the true number of MT instances at frame $t$ and $T$ is the total number of frames.
This choice of $M$ provides the opportunity for the counter function to learn about acceptable vs. non-acceptable performance for instance-level segmentation. To account for the overlapped area at a single frame, we use 3-D label tensor. Thus, our network learns about instances with conflicting areas that are either directly visible or hard to perceive due to occlusion. To compute the loss function we must evaluate the ground truth against our results. Since ground truth instances and segmentation results do not follow the same order, Hungarian algorithm is chosen as likely a solution to optimally match results with labels, using a cost matrix (Figure~\ref{Hung_alg}). 

We stipulate the cost function $f(\textbf{A},\textbf{B})$ as a measure of similarity between two binary masks $\textbf{A}$ and $\textbf{B}$ of the same size:
\begin{equation}
    f(\textbf{A},\textbf{B})=\frac{\sum\limits{(\textbf{A}\circ\textbf{B})}}{\sum\limits{(\textbf{A}+\textbf{B}-\textbf{A}\circ\textbf{B})}} ,
    \label{eq10}
\end{equation}
where $\textbf{A}\circ\textbf{B}$ represents the Hadamard product and summation is performed over all computed entries. Obtaining $f$ for each pair of a segmented instance and a ground truth, we from the cost matrix. In this matrix, Hungarian algorithm crosses out the higher values as absolute matches (such as $\textbf{Y}_{t}^{2}$ and $\textbf{I}_{t}^{1}$ in Figure~\ref{Hung_alg}) and optimize the rest of the matching procedure subsequently. As a result, we obtain a matrix where the $(i^{th},j^{th})$ element expresses the correspondence between the $i^{th}$ segmented instance (for any $i \in {1,...,{m}_{t}}$) and $j^{th}$ label, ($j \in {1,...,{n}_{t}}$) at time $t$.

\textit{\textbf{Loss function}}:
 For the loss function, we use the terms defined in Table~\ref{tab:table4} at time $t$, for the $i^{th}$ segmented instance and $j^{th}$ label instance.
 \begin{table}[!ht]
\caption{Definitions of terms in computing the loss function}
    \label{tab:table4}
    \begin{tabular}{!{\VRule[1pt]}c!{\VRule}l!{\VRule[1pt]}}
        \hline
        $\textbf{I}_{t}$  & current frame\\
        \hline
         $\textbf{I}^{\text{AR}_i}_{t}$& Predicted mask for attention region of instance $i$ \\
                 \hline
         $\textbf{I}^{i}_{t}$ & Predicted mask of segmented instance $i$ \\
                          \hline
         ${\text{S}^{i}}$  & Obtained score for segmenting instance $i$\\
         \hline
         $\textbf{Y}_{t}$ & 3-D matrix of binary masks at time $t$ \\
         \hline
         $\textbf{Y}^{\text{AR}_j}_{t}$ & Binary mask for attention region of instance $j$ \\
          \hline
         $\textbf{Y}^{j}_{t}$  & Binary mask of instance $j$\\
         \hline
         ${\text{S}^{j}}^{*}$  & True score\\
         \hline

    \end{tabular}
\end{table}

The total loss $L$ is defined as the sum of  $L_\text{att}$ (loss of the attention network), $L_\text{seg}$ (loss of the segmentation network) and $L_\text{count}$ (loss of the counter function): 
\begin{equation}
   L= L_\text{att}+L_\text{seg}+L_\text{count}.
    \label{eqL}
\end{equation}
Based on the matched results, we define $L_\text{att}$ as follow: 
\begin{equation}
    L_\text{att}(\textbf{I}_{t},\textbf{Y}_{t})=-\frac{1}{{m}_{t}}\sum\limits_{i,j}{{l_\text{att}}^{i,j}},
    \label{eq13}
\end{equation}
where
\begin{align}
   {l_\text{att}}^{i,j}=\begin{cases}
           f(\textbf{I}^{\text{AR}_i}_{t},\textbf{Y}^{\text{AR}_j}_{t}), & \text{if} \quad i \quad \text{matches} \quad j,\\
           0, & \text{otherwise.}
		\end{cases}
\label{eq11} 
\end{align}
\textcolor{black}{We use function $f$ as is defined in Equation \ref{eq10} to compute the number of shared pixels} between the $i^{th}$ proposed attention region and the $j^{th}$ true detection box as $l_\text{att}$. For segmentation loss, $L_\text{seg}$, we use 
\begin{equation}
    L_\text{seg}(\textbf{I}_{t},\textbf{Y}_{t})=-\frac{1}{{m}_{t}}\sum\limits_{i,j}{{l_\text{seg}}^{i,j}},
    \label{eq12}
\end{equation}
with ${l_\text{seg}}^{i,j}$ formulated to weight the similarity between $i^{th}$ segmented instance and the $j^{th}$ label instance and defined as
\begin{align}
   {l_\text{seg}}^{i,j}=\begin{cases}
           f(\textbf{I}^{i}_{t},\textbf{Y}^{j}_{t}), & \text{if} \quad i \quad \text{matches} \quad j,\\
           0, & \text{otherwise.}
		\end{cases}
\label{eq131} 
\end{align}
Finally, for $L_\text{count}$, we employ a monotonic score loss proposed by~\cite{ren2017end}, since counter function must compare high vs. low scores to make the network select more confident objects first:

\begin{align}
   l_\text{count}(\textbf{I}_{t},\textbf{Y}_{t})=&\frac{1}{M}\sum\limits_{i}-{{\text{s}^{i}}^*}\log{\Big(\min\limits_{u\leq i}\big({\text{s}^{i}}\big)\Big)}\nonumber \\
   &-(1-{\text{s}^{i}}^*)\log{\Big(1-\max\limits_{i\leq u}\big({\text{s}^{i}}\big)\Big)}
\label{eq132} 
\end{align}

During the test, our algorithm iterates over the test frame(s) and produces a score by the counter function. If this score falls under a certain threshold, the algorithm  stops iterating through the same frame, resets, and moves on to the next frame along the sequence of the time-lapse images.  

\subsection{Part 2: Data association}
\begin{figure}[!ht]
\centering
\includegraphics[height=.3\textwidth]{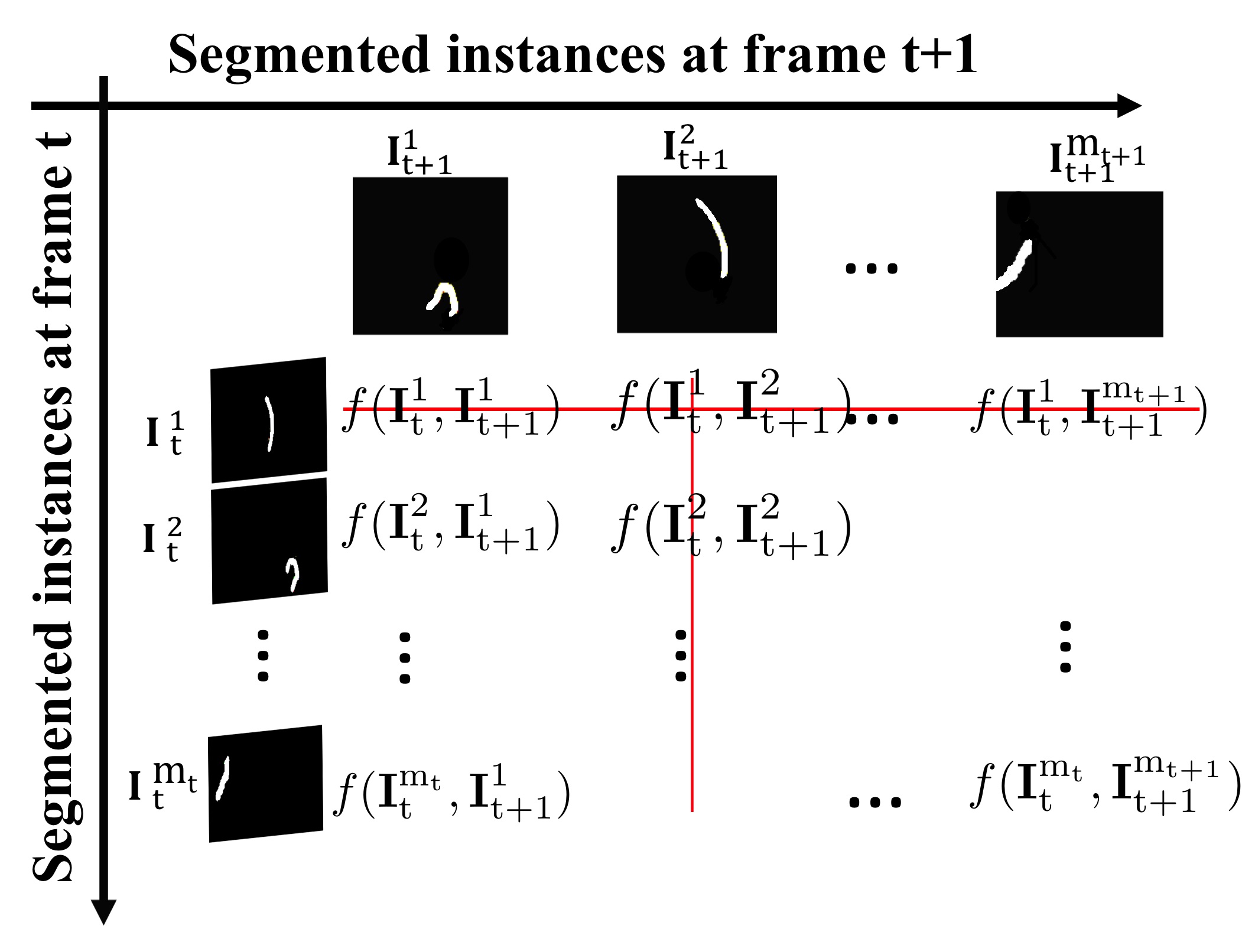}
\caption{The cost matrix of the Hungarian algorithm which measures the linking score between the segmented instances at frame $t$ and frame $t+1$(the higher values express more closeness).} 
\label{fig_hung2}
\end{figure}

After segmentation step, we associate the segmented instances for every two consecutive frames ($t$ and $t+1$). For this purpose, we use an associating-purpose Hungarian algorithm with a cost function $f(\textbf{I}_{t}^{i},\textbf{I}_{t+1}^{j})$ to represent the (${i}^{th}$,${j}^{th}$) element of the cost matrix (Figure~\ref{fig_hung2}). This function calculates the \textit{Intersection of Union} (IoU) between the ${i}^{th}$ segmented instance at time ${t}$ and the ${j}^{th}$ segmented instance at ${t+1}$. 
Doing so, we obtain three types of MT counts during the test:
\begin{itemize}
    \item $m_{t,t+1}\leq\min\big(m_{t},m_{t+1}\big)$ which is the number of instances being transferred to the next frame in a one-to-one manner.
    \item $m_{t,ext} = m_{t}-m_{t,t+1}$ indicating the number of instances at frame ${t}$ which left the scene by either sudden disappearance or exiting the frame.
    \item $m_{t,ent} = m_{t}-m_{t-1,t}$ expressing the number of instances at frame ${t}$ that enter the frame by a sudden appearance or simply move into the frame.
\end{itemize} 
These counts help to compute the displacement among segmented instances in successive frames.
\begin{figure*}
\centering
\includegraphics[width=1\textwidth]{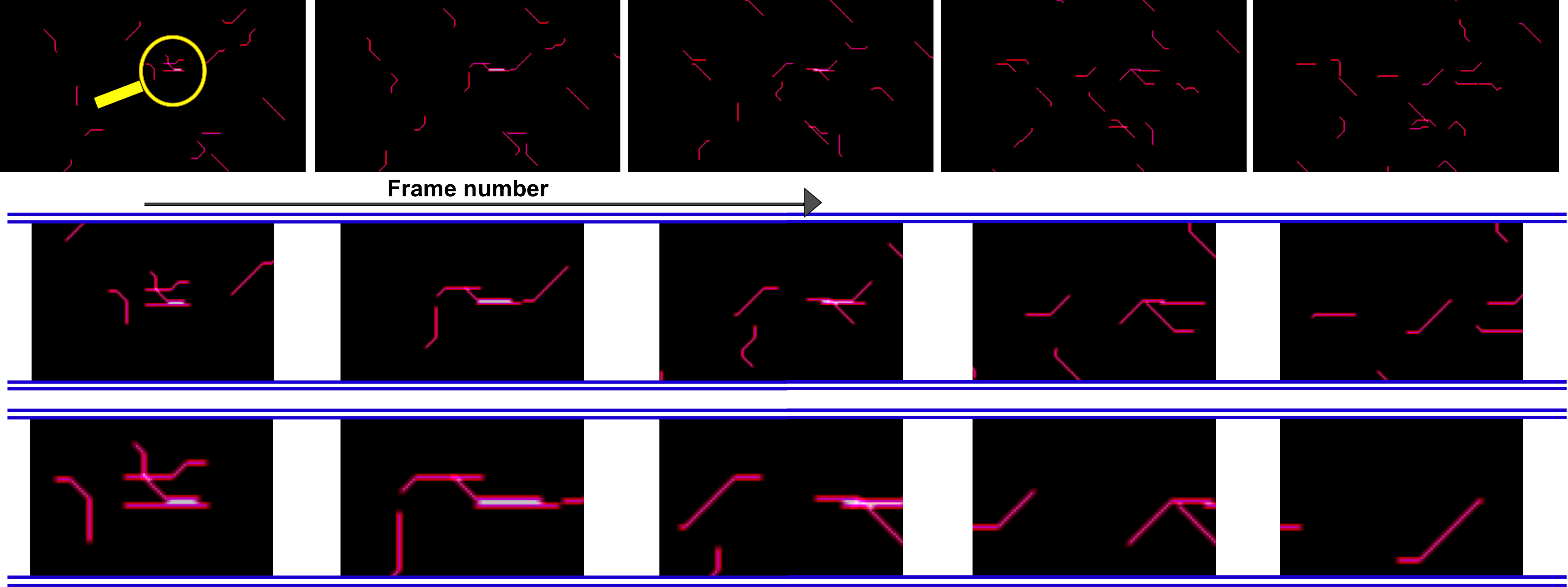}
\caption{left to right: temporally sub-sampled simulated data frames. First row includes original frames, second and third rows present their corresponding ($2.5\times2.5$) and ($5\times5$) zoomed versions (Color enhanced for better display).} 
\label{Simulateddata}
\end{figure*}
\begin{figure*}
\centering
\includegraphics[width=0.9\textwidth]{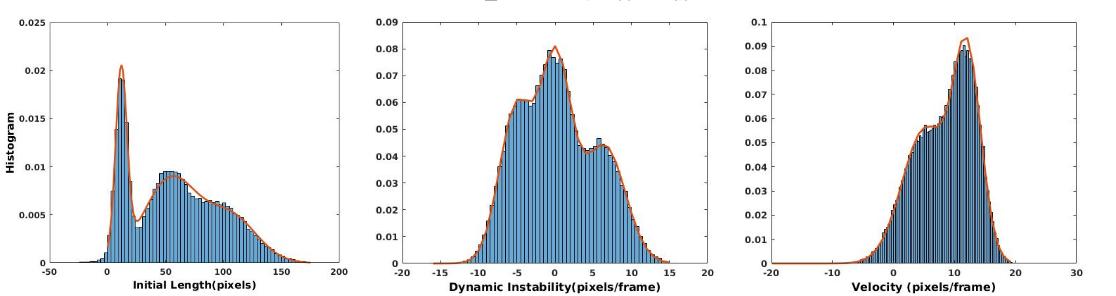}
\caption{Fitted PDF to the data obtained from 5000 individual MTs in real frames.} 
\label{Stat}
\end{figure*}
\section{Results}
\subsection{Data sets}
A common problem in biomedical imaging is the lack of large amounts of precisely annotated data. Herein, we have the same challenge; thus, we generate a type of data that closely simulates the actual microscopy images of MTs.
\subsubsection{Simulated data}
Similar to the real time-lapse images, the simulated data set should have RGB-channels captured from the central area of a large predefined frame to resemble the entrance and egress of MTs on the edges. Specific settings for generating a simulated sequence of the frames include: size and number of the frames, spatial and time resolution, initial number of the instances, their geometric specifications, and motion parameters. As it can be seen in Figure~\ref{Simulateddata}, MT instances are represented by \textit{wagon-train-shape} structures with identical width. To better imitate the statistical characteristics of the real MTs such as their length, velocity, and dynamic instability, we collect respective information from real imaging data. We take samples of real instances with at least two experts having unanimous labels for them. We then fit a multi-modal Gaussian distribution to each relevant characteristic of our samples, since it has the least mean square fitting error. This measure matches the maximum likelihood criteria, since a Gaussian distribution is known to happen.
The resulted distributions are shown in Figure~\ref{Stat}, where for each property, we set the number of modes in its distribution function equivalent to the number of states it may adopt. For instance, we use 3 modes in case of dynamic instability implying 3 phases of: shrinking, growing, and pause. In compilation of our simulated data, we make the MTs to take characteristics following the obtained distributions. Additionally, to mimic TIRF microscopy, another variable is taken into account indicating the sudden appearance/disappearance of the MTs. We utilize a contrast within our simulated image intensities where MTs have brighter interiors and even much lighter in their overlapping areas (see Figure~\ref{Simulateddata}). After generating the simulated data, we ignore the first few frames to avoid any bias induced by the initial conditions. All these details afford greater fidelity to the real-world microscopy images. \textcolor{black}{At the end, we generate 40 simulated image sequences, each has 379 frames of size $256\times256$ pixels. We use 4 randomly chosen sequences as the test data while the remainder is used for training with 5-fold cross validation setting.}
\subsubsection{Real data}
 \textcolor{black}{The real data is gathered in-house, containing 23 RGB, time-lapse sequences, each having a duration of $23.16\pm 12.68$ seconds. Every sequence is sampled at a rate of 16 frames per second with $256\times256$ pixels spatial dimension. We use 19 sequences of this data set for training repeated under cross-validation setting and utilize 4 other to test our algorithm.} This data set is annotated by three experts, who were asked to click on five points along the length from head to tail of \textit{what they interpret as a single MT}. This (five) was the smallest number found empirically to be sufficient extracting individual MTs in complex overlapping scenarios. The experts went through the whole time-lapse sequence to label MTs. Using these 5-point labels, we extract MT bodies with further processing based on thresholding, region growing, and template matching algorithms~\cite{Smasoudi2}. Since there are some intra-variations between the obtained labels, in each case we decided to use the most voted areas over all three labels as our ground truth. The resulting annotations are used to train our network. We also directly use the coordinates of the head of each MT in consecutive frames to have thorough description (${\boldsymbol{\delta}}_{GT}$) of the ground truth displacement. These vector labels are used in our final evaluation.

\subsection{Evaluation Metrics}
\textcolor{black}{To evaluate the segmentation performance of our proposed method in segmenting the $k^{th}$ instance at the $t^{th}$ frame, we employ conventional Jaccard index ($J$) as defined in Equation~\ref{BJC} :}
\begin{equation}
    \text{J}^{k}_{t}=\max_{j}\big({f(\textbf{I}^{k}_{t},\textbf{Y}^{j}_{t})}\big).
\label{BJC}    
\end{equation}
We report the best value for $J$ as well as the average $\text{J}^{k}_{t}$ value obtained over all instances in Table~\ref{tab:table6}. Since the ultimate goal in this study is to estimate MT motions along sequential frames, we quantify the overall performance of our proposed method in terms of displacement estimation. It should be noted that displacement is measured with respect to relocation of MT leading ends or heads. In the obtained trajectories, we subtract every two assigned instances at consecutive frames ($\textbf{I}_{t+1}-\textbf{I}_{t}$) to have the area presenting the head of the MT. We use the center of this area and present it using $(x,y)$. We define displacement ${\boldsymbol{\delta}}$ as in Eq.~\ref{eq135}:
\begin{equation}
    {\boldsymbol{\delta}}= [x_{t}\quad y_{t}\quad x_{t+1}\quad y_{t+1}]^T.
\label{eq135} 
\end{equation}
 \textcolor{black}{To evaluate the similarity between two displacement vectors: ${\boldsymbol{\delta}}$ and the ground truth (${\boldsymbol{\delta}}_{GT}$) in terms of their orientations \textit{or} magnitudes, we introduce a novel measure \textit{Vsim} in Eq.~\ref{eq_dice}}: 
\begin{equation}
    \textit{Vsim}({\boldsymbol{\delta}},{\boldsymbol{\delta}}_{GT})=\frac{{\boldsymbol{\delta}}.{\boldsymbol{\delta}}_{GT}}{\abs{{\boldsymbol{\delta}}}^2+{\abs{{\boldsymbol{\delta}}_{GT}}^2}}+\frac{1}{2}.
    \label{eq_dice}
\end{equation}

We obtain the best \textit{Vsim} value (i.e., BVs) for the $i^{th}$ displacement vector obtained at transition from $t^{th}$ to the ${t+1}^{th}$ frame,  and reported $\text{BVs}$ and average $\text{BVs}_{\,t}^{\,i}$  over all displacement vectors:
\begin{align}
\text{BVs}_{\,t}^{\,i}=\max_{\boldsymbol{\delta}_{GT}}\big(\textit{Vsim}({\boldsymbol{\delta}^{\,i}},{\boldsymbol{\delta}}_{GT})\big),\quad i \in \{1,...,m_{t,t+1}\}.
    \label{eq_BVs}
\end{align}
 \begin{figure*}[t]
\centering
\includegraphics[width=0.9\textwidth]{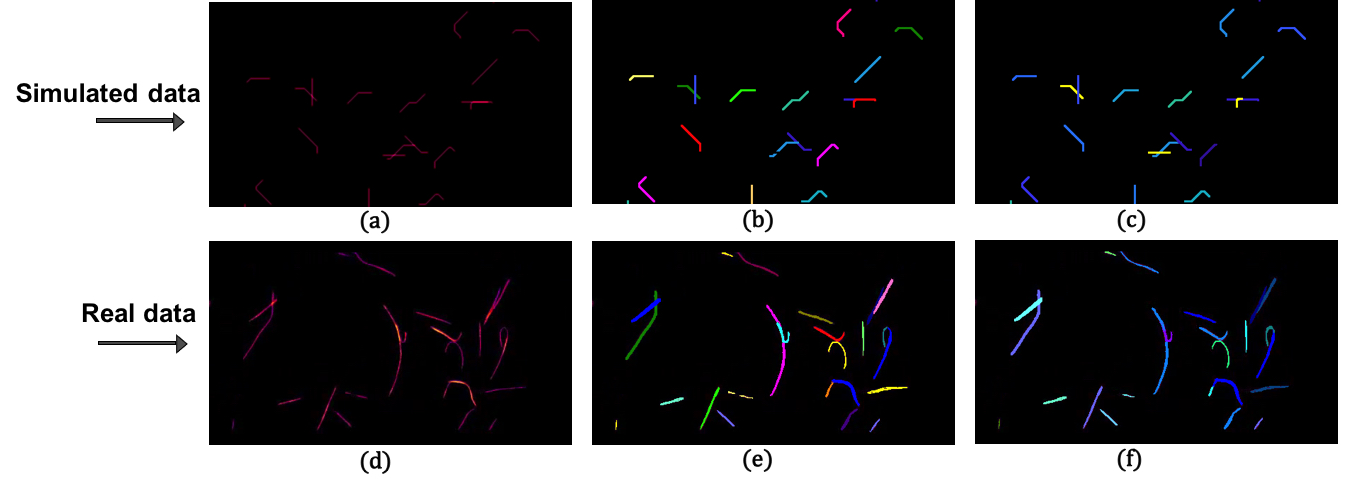}
\caption{Instance-level segmentation results, for simulated data (first row) and real data (second row). Original frame (a), ground truth annotations (b) and the results of the proposed method (c) are shown.} 
\label{Results}
\end{figure*}
\begin{table*}[t]
  \begin{center} 
    \caption{\textcolor{black}{Instance-level MT segmentation performance in terms of {best Jaccard similarity coefficient} (\text{\textbf{J}}), {false positive rate} (\text{\textbf{FPR}}), and {false negative rate} (\text{\textbf{FNR}})}}
    \label{tab:table6}
    \begin{tabular}{l|c|cccccccc} 
        &\textbf{Dataset} &\textbf{J}  & \textbf{FPR}& \textbf{FNR}\\
      \hline
       Adaptive Template Matching~\cite{Smasoudi2}& Real& \textcolor{black}{{0.533}} &\textcolor{black}{{0.596}}&\textcolor{black}{{0.415}} \\
       PMM Kalman Smoother~\cite{Smasoudi2}& Real&\textcolor{black}{{0.474}} &\textcolor{black}{\textbf{0.343}}&\textcolor{black}{{0.285}}\\
      Ours (OF, L=5)& Real&\textcolor{black}{\textbf{0.681}} &\textcolor{black}{{0.455}}&\textcolor{black}{\textbf{0.219}}\\

    \end{tabular}
  \end{center}
\end{table*}

\begin{table*}[t]
  \begin{center} 
    \caption{Instance-level MT velocity estimation performance in terms of {best Vsim} (\text{\textbf{BVs}}), {false discovery rate} (\text{\textbf{FDR}}), {false negative rate} (\text{\textbf{FNR}}), and {Difference in Counting} (\text{\textbf{DiC}}).}
    \label{tab:table3}
    \begin{tabular}{l|c|cccccccc} 
        &\textbf{Dataset} &\textbf{BVs}  & \textbf{FDR}& \textbf{FNR} & \textbf{DiC}&$
\mathbf{\text{DIC}_\text{ext}}$&$\mathbf{\text{DIC}_\text{ent}}$\\
      \hline
       Adaptive Template Matching~\cite{Smasoudi2}& Real&  0.293 &0.762 &0.678&0.431&0.624&0.829\\
       PMM Kalman Smoother~\cite{Smasoudi2}& Real&0.568 &0.372 &0.391 &0.512&0.438&0.329 \\
      Ours (OF, L=5)& Real& \textbf{0.632} &\textbf{0.237} &\textbf{0.287} &\textbf{0.116}&\textbf{0.363}&\textbf{0.313} \\

    \end{tabular}
  \end{center}
\end{table*}

After assigning each displacement to its equivalent ground truth, we count the true positives. We define true positive according to the intra-variation existing between the three experts labeling outcomes, where we let our network to make errors less than the difference between the labels obtained from the experts. \textcolor{black}{\textit{False discovery rate} (FDR) is measured as a ratio of the false positives to the total number of computed displacements at each frame where false positives are the vectors that were not assigned to any ground truth or if assigned, they did not fulfill the requirements of being a true positive. We also define \textit{false negative rate} (FNR) as the ratio of the ground truth vectors with no attributed estimated vector to the total number of the ground truth vectors at each frame.} Eventually, \textit{Difference in Counting} (DiC) is used to compare the counted number of segmented instances against the ground truths:
 
\begin{align}
    \abs{DiC}_{trans}=\frac{1}{T}\sum_{t}{\frac{\abs{m_{t,t+1}-n_{t,t+1}}}{n_{t,t+1}}}, 
     \nonumber\\
  \abs{DiC}_{ext}=\frac{1}{T}\sum_{t}{\frac{\abs{m_{t,ext}-n_{t,ext}}}{n_{t,ext}}},\label{eq_dif_count}\\
    \abs{DiC}_{ent}=\frac{1}{T}\sum_{t}{\frac{\abs{m_{t,ent}-n_{t,ent}}}{n_{t,ent}}},\nonumber
\end{align}
where sub-indexes $trans$, $ext$, and $ent$ respectively denote the MTs which transitioned, exited, and entered the $t^{th}$ frame.
\subsection{Qualitative evaluations}
Some of instance-level MT segmentation results are presented in Figure~\ref{Results}. As shown, there is a significant positive correlation between the ground truth and results of our best model for both simulated and the real data. The visually perceivable results of MT tracking are provided in Appendix \ref{FirstAppendix}, where MTs' heads displacement are demonstrated with their velocity amplitude.
\subsection{Quantitative evaluations}
We analyzed the performance of our algorithm in more details by providing \textcolor{black}{five} Tables. 
All values are obtained using threshold value of 0.23 for counter function and are averaged over all frames, all instances (if applicable) in test data. This threshold value minimizes the average $FN \times FP$ (in terms of segmentation results) for a set of 100 randomly chosen frames from the training set.  \textcolor{black}{Having such configuration, our optimum design runs at 250 ms per frame ($256\times256$) to perform segmentation.}

\textcolor{black}{Tables~\ref{tab:table6} and\textcolor{black}{~\ref{tab:table3}} illustrate the distinction of our method (optimum design) in terms of segmentation and velocity estimation against two baseline methods: \textit{adaptive template matching} and \textit{piece-wise stationary multiple motion Kalman smoother (PMM Kalman smoother)}}. Adaptive template matching updates an initial set of templates with results obtained from 3 past frames~\cite{Smasoudi2}. PMM Kalman smoother uses piece-wise stationary multiple motion model. As shown quantitatively in both tables, our framework outperforms the baseline results. A reduction of at least 0.235 in FDR and 0.104 in FNR in velocity estimation confirm the greater capability of our method in dealing with complicated problem of instance-level MTs segmentation and tracking. \textcolor{black}{Results express a reduction of at least 0.066 in FNR, along with an acceptable FPR result.} This value of the FPR does not degrade the performance of our algorithm in velocity estimation. \textcolor{black}{Additionally, we have compared our algorithm to two instance-level segmentation methods (~\cite{Bai2017} and~\cite{Romera-Paredes2016}). Confirmed by Table~\ref{table_compare}, our algorithm results in significant reduction of the FNR.}

Table~\ref{tab:table5} demonstrates the potentials of using RNN component comparing to CNN component as a visually attentive operator. Tables~\ref{tab:table1} and~\ref{tab:table2} summarize the contrast between two possibilities of using original frames or their respective OFs as part of the input to visual attention. Results show that using OF can increase the precision up to 0.29 in case of real data.

\textit{\textbf{Ablation study:}}
\textcolor{black}{To assess our proposed visual attention module, we substitute the CNN+RNN with CNN in Exp-1. In case of using CNN only, the descriptive characteristics of the bounding box are produced by the last fully connected layer. This experiment investigates our proposed design against methods  in~\cite{he2017fine,fu2017look,peng2018object}, which results are presented in Table~\ref{tab:table5}.} \textcolor{black}{ As shown, The spatial reasoning of LSTM  leverages its ability in comparison to using CNNs. Although, results obtained in either cases are of a comparable order.}
In another experiment (Exp-2), we demonstrate how the injection of temporal information into our instance-level segmentation network improves the quality of displacement estimation. To this end, we use various number of frames to supply the network. Results, reported in Tables~\ref{tab:table1} and~\ref{tab:table2}, support the idea that using neighboring frames leads to better estimation and less miss-detections (false negatives). 
 
In Exp-3, we examine whether temporal information affords better estimation in form of raw neighboring frames or their respective OF. Comparing Tables~\ref{tab:table1} and~\ref{tab:table2} reveals the results of this experiment, indicating a dramatic progress in case of OF. It is because using OF adds an initial motion clue to information at the current frame, which leads to more accurate detection path. 

 Finally, Exp-4 is designed to study the network performance in case of simulated vs. real data. Results from Tables~\ref{tab:table1} and~\ref{tab:table2} express that real data is harder to analyze. For both data categories, we get improved results from our trained model compared to the baseline methods. Unlike our all-embracing labels for the simulated data (which includes the exact pixels of every MT), we had only 5-marker coordinates directly annotated by experts due to limited time and expertise. While the questionable nature of these labels can crucially resolve the network's performance in segmentation, it could not prevent a significant contribution to displacement estimation. 
 
 \textcolor{black}{Exp-5 and Exp-6 are aimed to study the robustness of our algorithm. In Exp-5, we quantify the algorithm performance in terms of instance-level MT segmentation for different crowdedness in each frame. As it is demonstrated by Table~\ref{tab:table7}, overpopulated scenes degrade the segmentation results in terms of significantly higher FPR and FNR. However, such failure rate specifically manifests itself when the algorithm faces a crowd of more than 30 MTs whitin a frame. In Exp-6, the quality of velocity estimation is evaluated against sampling rates of a time-lapse sequence. According to Table~\ref{tab:table8}, while downsampling (time-wise) deteriorates the performance in an obvious way, sampling rate of 2 and 4 lead to unacceptable results. }


\begin{table*}[ht!]
  \begin{center} 
    \caption{Instance-level MT velocity estimation using raw frames (threshold=0.23, L = temporal window length).}
    \label{tab:table1}
    \begin{tabular}{l|c|cccccccc} 
        &\textbf{Dataset} &\textbf{BVs}  & \textbf{FDR}& \textbf{FNR} & \textbf{DiC}&$
\mathbf{\text{DIC}_\text{ext}}$&$\mathbf{\text{DIC}_\text{ent}}$\\
      \hline
       L=1& Sim&  0.320 &0.420 &0.230 &0.356&0.611&\textbf{0.278}\\
       L=3& Sim&  0.465 &0.120 &0.313 &0.273&0.454&0.438\\
       L=5& Sim& \textbf{0.665} &\textbf{0.001} &\textbf{0.149}&\textbf{0.118}&\textbf{0.413}&0.404\\\hline
       L=1& Real&  0.249 &0.215 &0.320 &0.563&0.682&0.439\\
       L=3& Real&0.544 &0.327 &0.319&0.542&0.512&0.418 \\
       L=5& Real&  \textbf{0.583} &\textbf{0.321} &\textbf{0.314} &\textbf{0.532}&\textbf{0.463}&\textbf{0.374}\\

    \end{tabular}
  \end{center}
\end{table*}

\begin{table*}[ht!]
  \begin{center}
    \caption{instance-level MT velocity estimation using OF, (threshold=0.23, L = temporal window length).}
    \label{tab:table2}
      \begin{tabular}{l|c|cccccccc} 
        &\textbf{Dataset} &\textbf{BVs}  & \textbf{FDR}& \textbf{FNR}  & \textbf{DIC}&$\mathbf{\text{DIC}_\text{ext}}$&$\mathbf{\text{DIC}_\text{ent}}$\\
      \hline
       L=3& Sim& 0.705 &0.023 &0.156 &0.086&0.319&0.237 \\
       L=5& Sim& \textbf{0.712} &\textbf{0.001} &\textbf{0.083} &\textbf{0.081}&\textbf{0.211}&\textbf{0.186} \\\hline
       L=3& Real& 0.588 &0.268 &0.151 &0.171 &0.390&0.349  \\
       L=5& Real& \textbf{0.632} &\textbf{0.237} &\textbf{0.287}&\textbf{0.116}&\textbf{0.363}&\textbf{0.313}  \\
    \end{tabular}
  \end{center}
\end{table*}

 \begin{table*}[ht!]
  \begin{center} 
    \caption{Instance-level MT velocity estimation using OF, for different potential architectures of the visual attention module\ (threshold=0.23, L=5).}
    \label{tab:table5}
    \begin{tabular}{l|c|cccccccc} 
        &\textbf{Dataset} &\textbf{BVs}  & \textbf{FDR}& \textbf{FNR}& \textbf{DiC}&$
\mathbf{\text{DIC}_\text{ext}}$&$\mathbf{\text{DIC}_\text{ent}}$\\
      \hline
        \textcolor{black}{CNN1 (10-layers)}& \textcolor{black}{Sim}& \textcolor{black}{{0.556}} &\textcolor{black}{{0.159}}&\textcolor{black}{{0.318}} &\textcolor{black}{{0.217}} &\textcolor{black}{{0.511}} &\textcolor{black}{{0.526}}\\
        \textcolor{black}{CNN1 (15-layers)}& \textcolor{black}{Sim}& \textcolor{black}{{0.564}} &\textcolor{black}{{0.154}}&\textcolor{black}{{0.303}} &\textcolor{black}{{0.221}} &\textcolor{black}{{0.513}} &\textcolor{black}{{0.515}}\\
       \textcolor{black}{CNN (8-layers)+LSTM}& \textcolor{black}{Sim}& \textbf{0.712} &\textbf{0.001} &\textbf{0.083} &\textbf{0.081}&\textbf{0.211}&\textbf{0.186} \\\hline
        \textcolor{black}{CNN1 (10-layers)}& \textcolor{black}{Real}& \textcolor{black}{{0.412}} &\textcolor{black}{{0.347}}&\textcolor{black}{{0.452}} &\textcolor{black}{{0.560}} &\textcolor{black}{{0.487}} &\textcolor{black}{{0.432}}\\
        \textcolor{black}{CNN1 (15-layers)}& \textcolor{black}{Real}& \textcolor{black}{{0.418}} &\textcolor{black}{{0.344}}&\textcolor{black}{{0.439}} &\textcolor{black}{{0.551}} &\textcolor{black}{{0.479}} &\textcolor{black}{{0.415}}\\
       \textcolor{black}{CNN (8-layers)+LSTM}& \textcolor{black}{Real}&\textbf{0.632} &\textbf{0.237} &\textbf{0.287}&\textbf{0.116}&\textbf{0.363}&\textbf{0.313} \\

    \end{tabular}
  \end{center}
\end{table*}
  
\begin{table*}[t]
  \begin{center} 
       \caption{\textcolor{black}{Average performance of instance-level MT segmentation over 50 frames simulated to contain 30 number of MT instances.}}
        \label{table_compare}
    \begin{tabular}{cccc} 
       Method &\textbf{BJc}  & \textbf{FPR}& \textbf{FNR}\\
      \hline
       Recurrent Instance Segmentation~\cite{Romera-Paredes2016} &\textcolor{black}{{0.651}}&\textcolor{black}{{0.097}}&\textcolor{black}{{0.311}}\\
      Deep Watershed Transform for Instance Segmentation~\cite{Bai2017}&\textcolor{black}{{0.514}} &\textcolor{black}{{0.056}}&\textcolor{black}{{0.402}}\\
      \textbf{Ours (OF, L=5)} &\textbf{\textcolor{black}{{0.743}}} &\textbf{\textcolor{black}{{0.028}}}&\textbf{\textcolor{black}{{0.069}}}
    \end{tabular}
  \end{center}
\end{table*}

\begin{table*}[!ht]
%
\parbox{.45\linewidth}{
      \centering
      \caption{\textcolor{black}{Average performance of instance-level MT segmentation (OF, L=5) over 50 frames simulated to contain certain number of MT instances.}}
    \begin{tabular}{cccc} 
       MT number  &\textbf{BJc}  & \textbf{FPR}& \textbf{FNR}\\
      \hline
       10 &\textcolor{black}{{0.825}}&\textcolor{black}{{0.012}}&\textcolor{black}{{0.046}}\\
      20 &\textcolor{black}{{0.796}} &\textcolor{black}{{0.018}}&\textcolor{black}{{0.065}}\\
        30 &\textcolor{black}{{0.743}} &\textcolor{black}{{0.028}}&\textcolor{black}{{0.069}}\\
      40 &\textcolor{black}{{0.711}} &\textcolor{black}{{0.055}}&\textcolor{black}{{0.179}}\\
    \end{tabular}
%

    \label{tab:table7}}
    $\qquad$
 \parbox{.5\linewidth}{
 \centering 
    \caption{\textcolor{black}{Average performance of instance-level MT velocity estimation (OF) over 10 sequences of length 4 seconds from the real data sub-sampled to have certain frame rates (fps).}}

    \begin{tabular}{l|c|cccccccc} 
        &\textbf{BVs}  & \textbf{FDR}& \textbf{FNR} & \textbf{DiC}&$
\mathbf{\text{DIC}_\text{ext}}$&$\mathbf{\text{DIC}_\text{ent}}$\\
      \hline

      \textbf{Frame rate=16} & \textbf{\textcolor{black}{0.632}} &\textbf{\textcolor{black}{0.237}} &\textbf{\textcolor{black}{0.287}}&\textbf{\textcolor{black}{0.116}}&\textbf{\textcolor{black}{0.363}}&\textbf{\textcolor{black}{0.313}} \\
      Frame rate=8 &\textcolor{black}{{0.613}} &\textcolor{black}{{0.259}}&\textcolor{black}{{0.288}}&\textcolor{black}{{0.119}} &\textcolor{black}{{0.367}}&\textcolor{black}{{0.319}}\\
      Frame rate=4 &\textcolor{black}{{0.506}} &\textcolor{black}{{0.383}}&\textcolor{black}{{0.315}}&\textcolor{black}{{0.377}} &\textcolor{black}{{0.441}}&\textcolor{black}{{0.423}}

    \end{tabular} 
     \label{tab:table8}}   
\end{table*}

\section{Discussions and Concluding Remarks}
Several methods have been proposed to \textcolor{black}{facilitate automated tracking of the growing ends of MTs. However, there is a paucity of automated approaches available to extract and measure the velocities of MTs in \textit{in-vitro} gliding assays. The major hurdle is having the frequent MT-MT interactions causing abrupt changes in MT motion trajectories}. The nature of MT motion in these assays \textcolor{black}{renders manual inspection and simple modeling tools inadequate for velocity characterization and measurement.} Both the human eye and simple methods tend to be biased by local changes in population density. The ever changing patterns of MTs motion make this characterization even more challenging. In summary, these limitations necessitate an automated approach with higher accuracy to accelerate the process of segmentation, tracking, and analysis. 

In this study, we employed new algorithms  resulting in fewer false positives and false negatives, while accounting for motion complexity. Our proposed approach iterates through attention and segmentation blocks to recognize one instance at a time. The presented attention network mimics human vision to set attention boundaries. Once attention network finds candidate regions of MTs, a back-to-back encoder-decoder engine is used to segment the relevant instances inside the candidate regions. 
 
Despite achieving the state-of-the-art performance in MT segmentation and velocity estimation using our designed network, there are still areas for potential improvement in near future. Expanding the repertoire of real data sets, in the form of an extended library of time-lapse image series, is one of them. For simulated data sets, increasing the data size with realistic augmentation strategies may leverage the training quality. In this regard, conditional generative adversarial networks are potential methods to apply~\cite{chuquicusma2018fool}. \textcolor{black}{Previously, authors adopted ``attention" to get fine-grained details of a single object in an image. This conventional attention concept is an artificial version of human vision in ``looking at a particular scene while giving deep attention into details of a small compartment in the same scene". However, in our work, we adopted the ``attention" to exploit the spatial relation that exists among different instances of the MTs in a single frame.  Hence, this version of attention can be interpreted as eye movement to switch attention from one instance to another, while they are not directly related.} Our algorithm concentrates on different regions of a single frame and segments an instance in each region. 
Once all the results from individual frames are available, our algorithm associates the results to extract trajectories. \textcolor{black}{We believe our algorithm has the least cost to perform instance-level MT segmentation and tracking among all other available methods. For instance, while the attention-based work by~\cite{vaswani2017attention} can be a great fit for 1-D machine translation applications, its direct extension to our 2-D+t problem is costly. The intrinsic elements of this method such as query, keys, and values can not simply be mapped with a dot product.} This structure can be extended into a version that could simultaneously incorporate both temporal and spatial dimensions of this problem. Such structure bypasses the requirement for a following data association algorithm and generates results with additional spatial accuracy and temporal smoothness. \textcolor{black}{As this research progresses, it is hoped that it will provide an automatic segmentation platform to help researchers to better study the molecular basis for the motor-dependent spatial organization of MTs in both interphase and mitotic cells.}

\appendix

\begin{figure}[!ht] 
 \centering
\subfloat[]{\includegraphics[width=.2\textwidth]{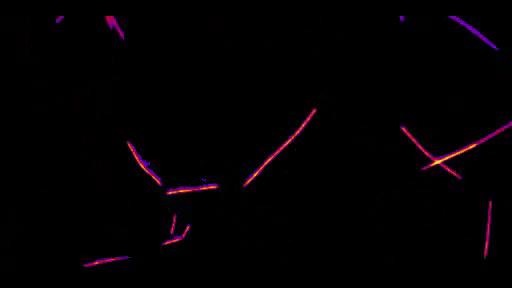}}
\hspace{.15cm}
\subfloat[]{\includegraphics[width=.2\textwidth]{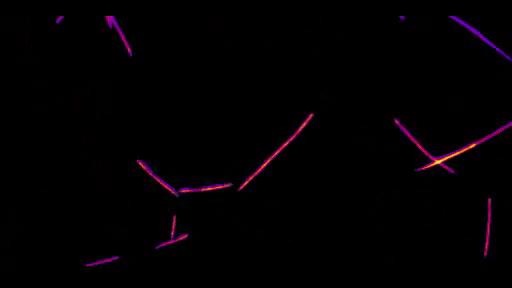}}
\\
\subfloat[]{\includegraphics[width=.2\textwidth]{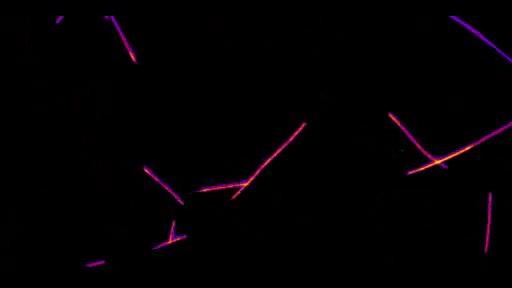}}
\hspace{.15cm}
\subfloat[]{\includegraphics[width=0.2\textwidth]{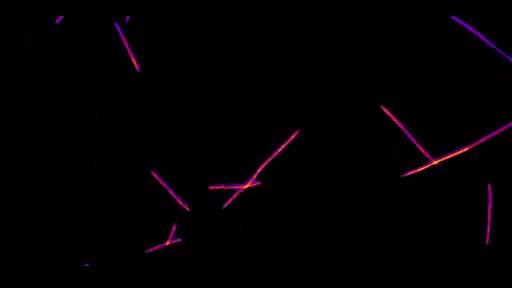}} 
\\
\subfloat[]{\includegraphics[width=.2\textwidth]{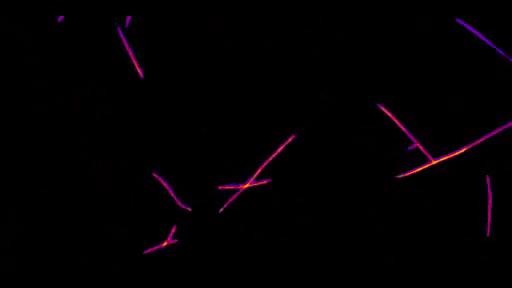}}
\hspace{.15cm}
\subfloat[]{\includegraphics[width=.2\textwidth]{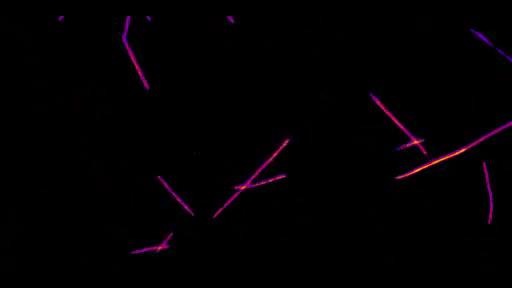}}
\\
\subfloat[]{\includegraphics[width=.2\textwidth]{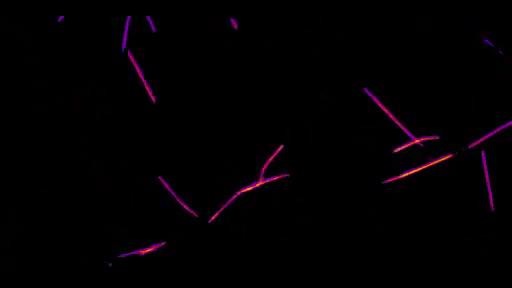}}
\hspace{.15cm}
\subfloat[]{\includegraphics[width=0.2\textwidth]{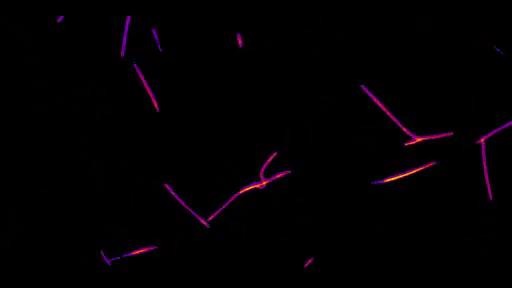}} 
\\
\subfloat[]{\includegraphics[width=.2\textwidth]{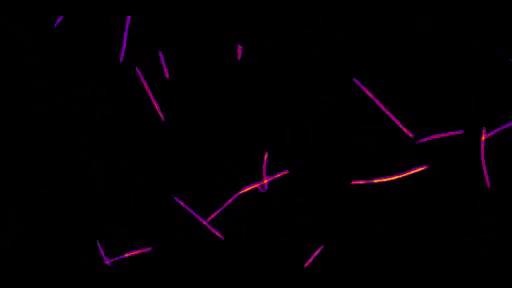}}
\hspace{.15cm}
\subfloat[]{\includegraphics[width=.2\textwidth]{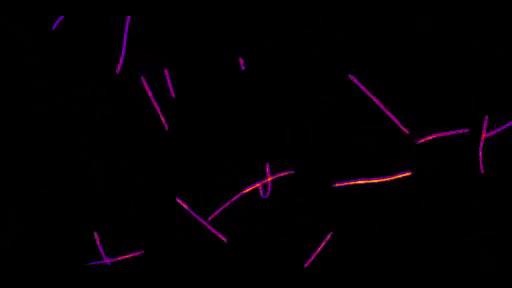}}
\\
\subfloat[]{\includegraphics[width=.2\textwidth]{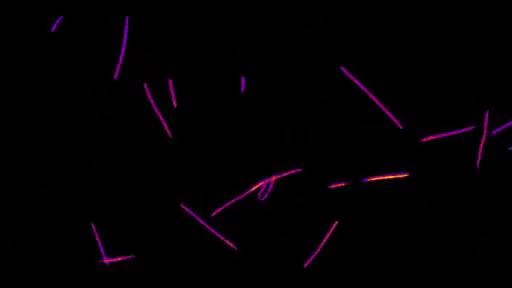}}
\hspace{.15cm}
\subfloat[]{\includegraphics[width=0.2\textwidth]{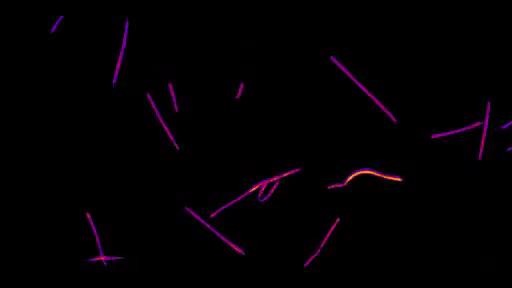}} 
\\
\subfloat[]{\includegraphics[width=.2\textwidth]{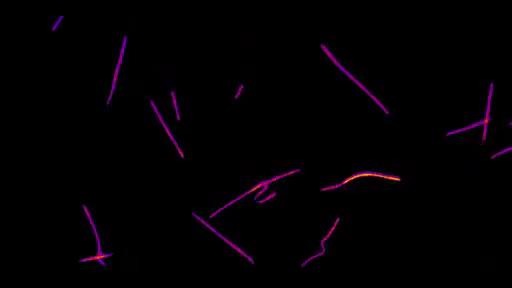}}
\hspace{.15cm}
\subfloat[]{\includegraphics[width=.2\textwidth]{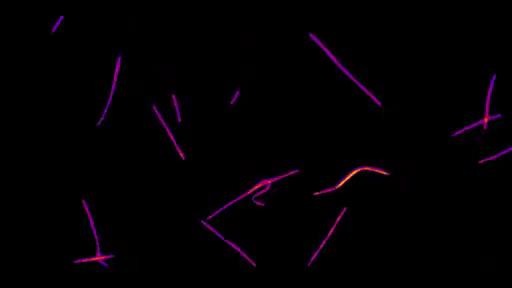}}
\caption{Extracted frames from a sample time-lapse sequence with the rate of 4fps}
\label{Fig_app1}
\end{figure}

\begin{figure}[!ht] 
\subfloat[]{\includegraphics[width=0.2\textwidth]{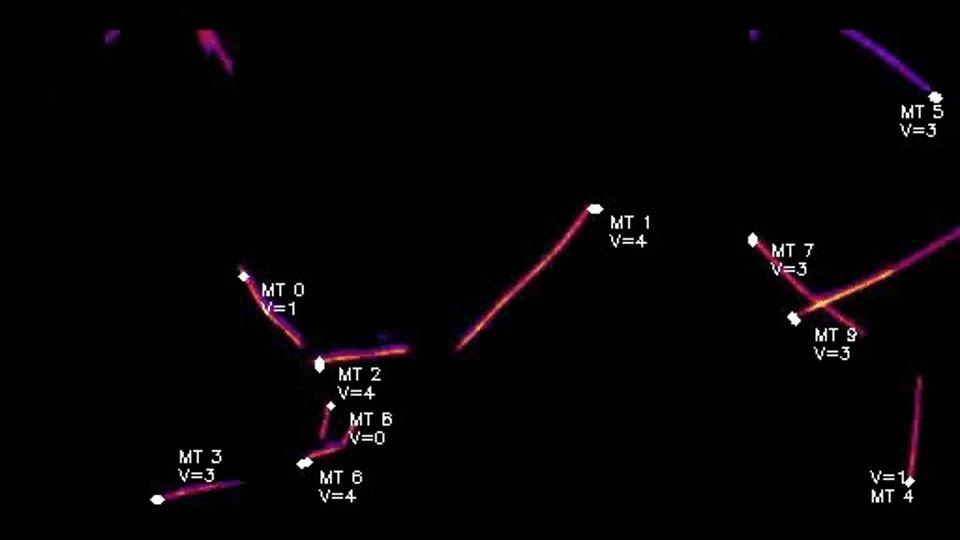}}
\hspace{.15cm}
\subfloat[]{\includegraphics[width=0.2\textwidth]{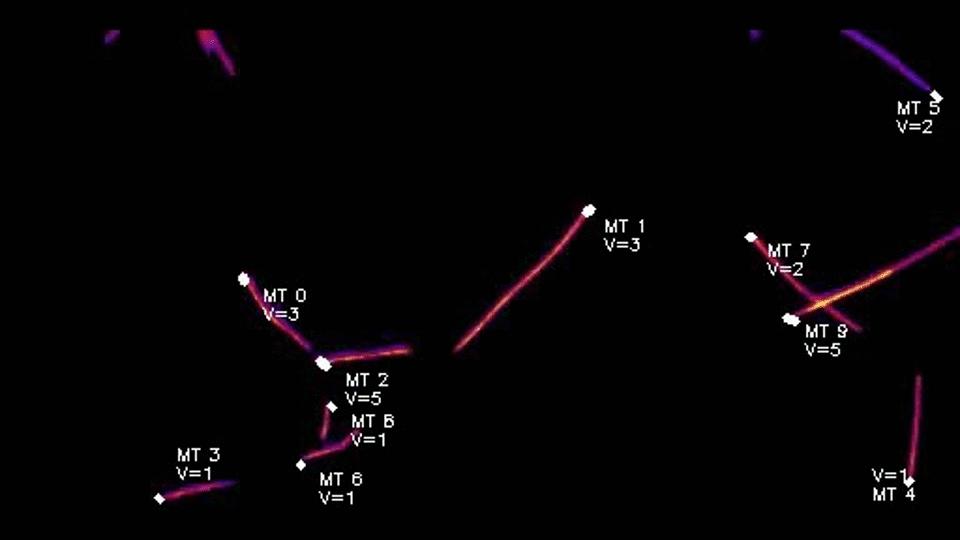}}
\\
\subfloat[]{\includegraphics[width=0.2\textwidth]{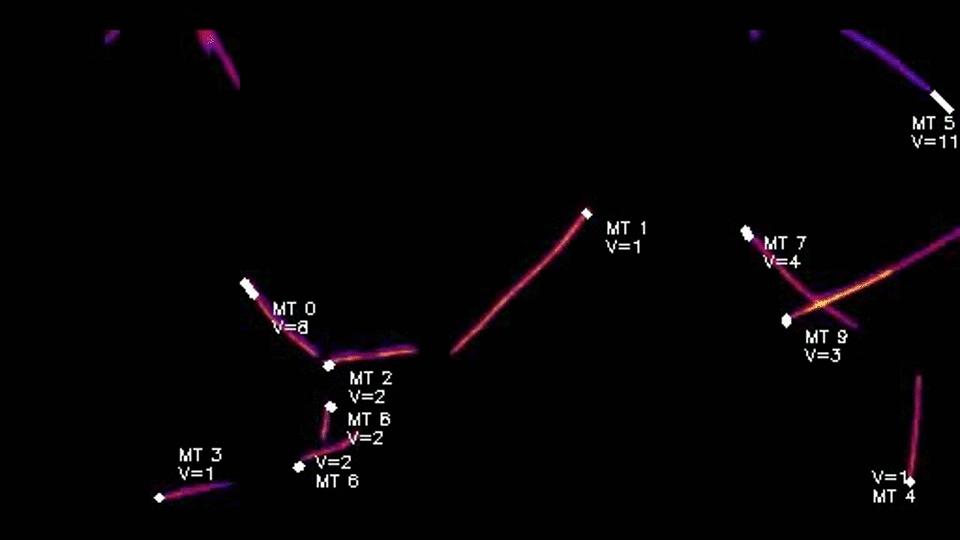}}
\hspace{.15cm}
\subfloat[]{\includegraphics[width=0.2\textwidth]{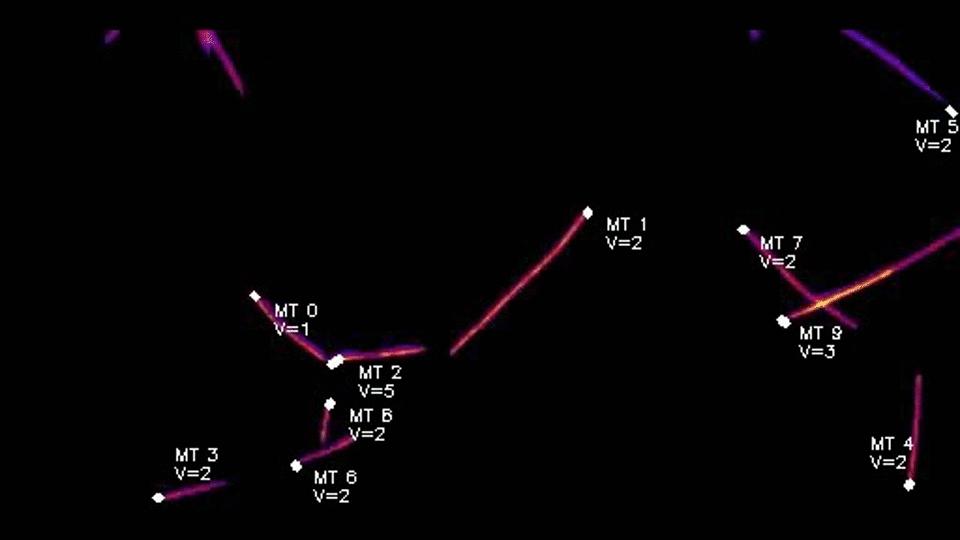}} 
\\
\subfloat[]{\includegraphics[width=0.2\textwidth]{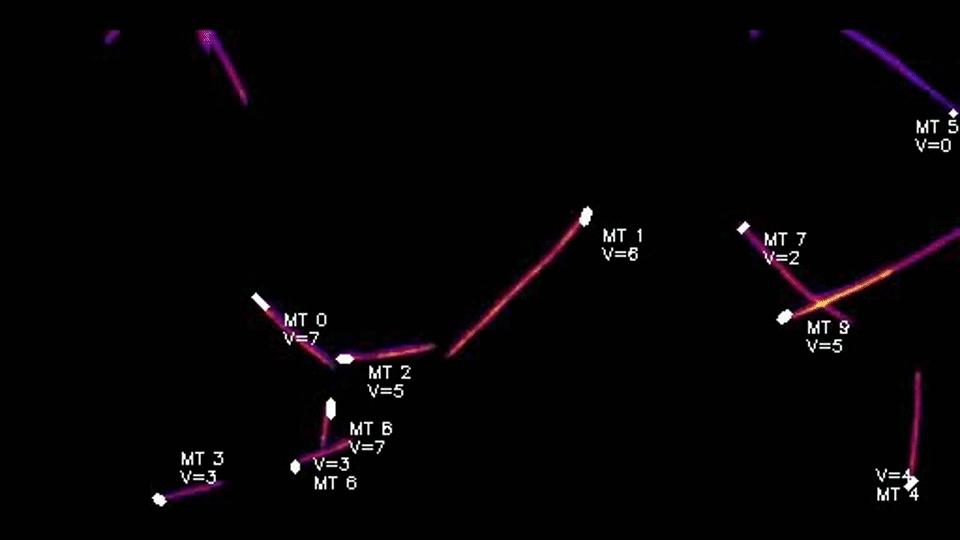}}
\hspace{.15cm}
\subfloat[]{\includegraphics[width=0.2\textwidth]{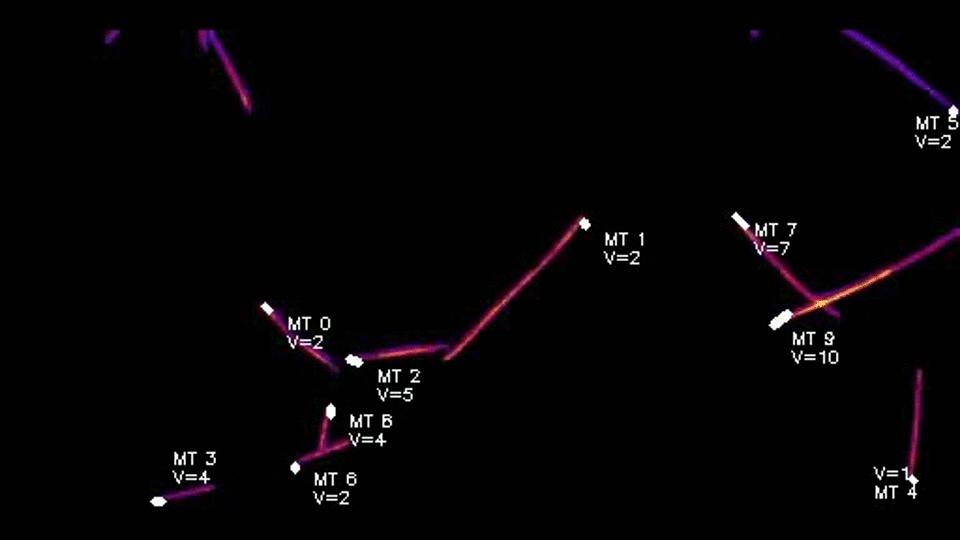}}
\\
\subfloat[]{\includegraphics[width=0.2\textwidth]{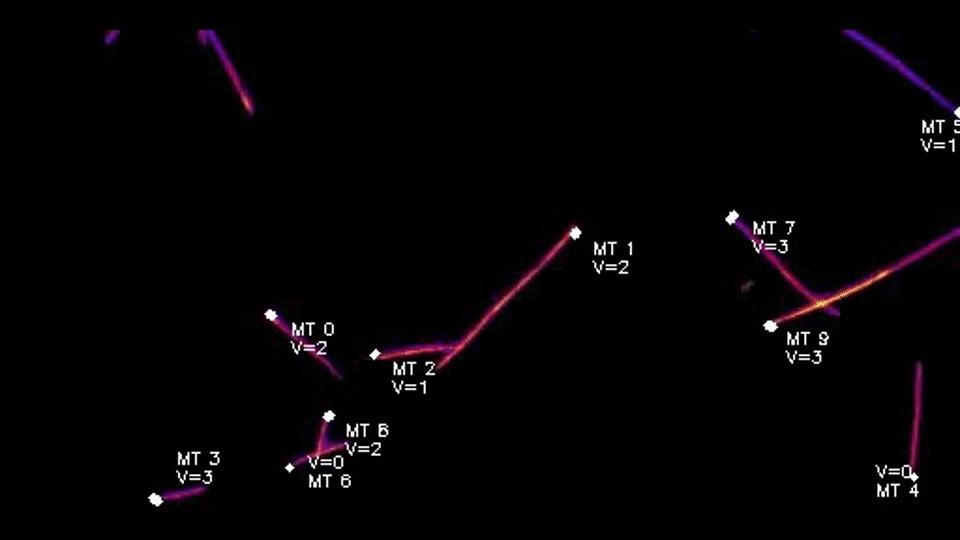}}
\hspace{.15cm}
\subfloat[]{\includegraphics[width=0.2\textwidth]{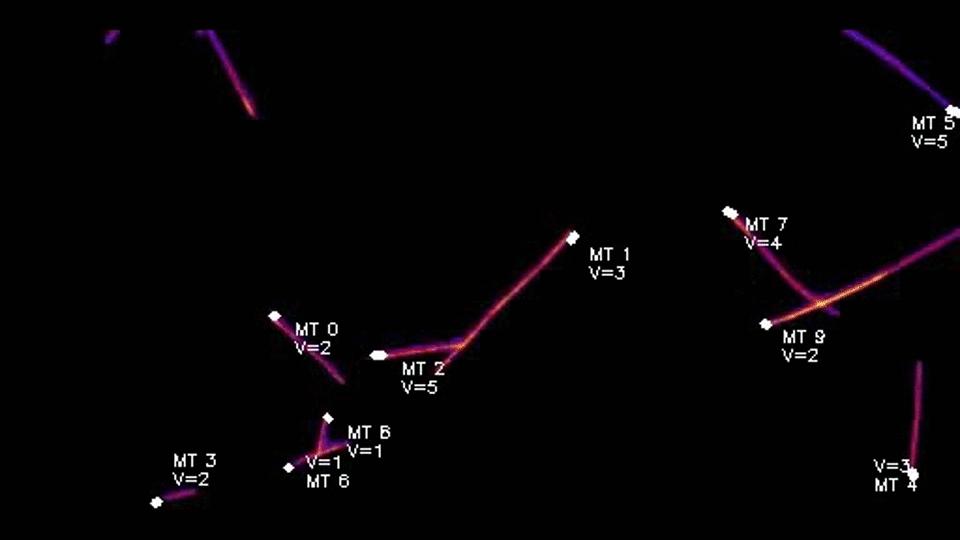}} 
\\
\subfloat[]{\includegraphics[width=0.2\textwidth]{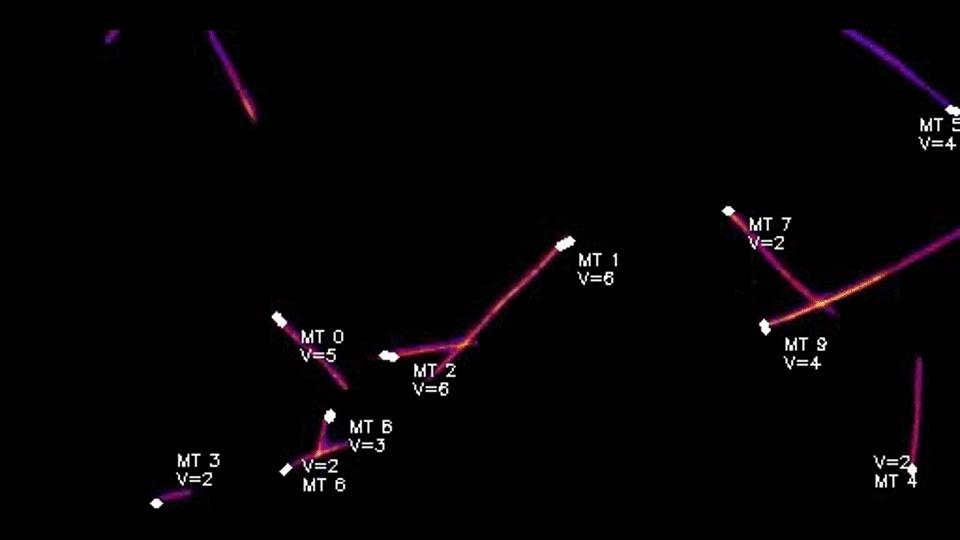}}
\hspace{.15cm}
\subfloat[]{\includegraphics[width=0.2\textwidth]{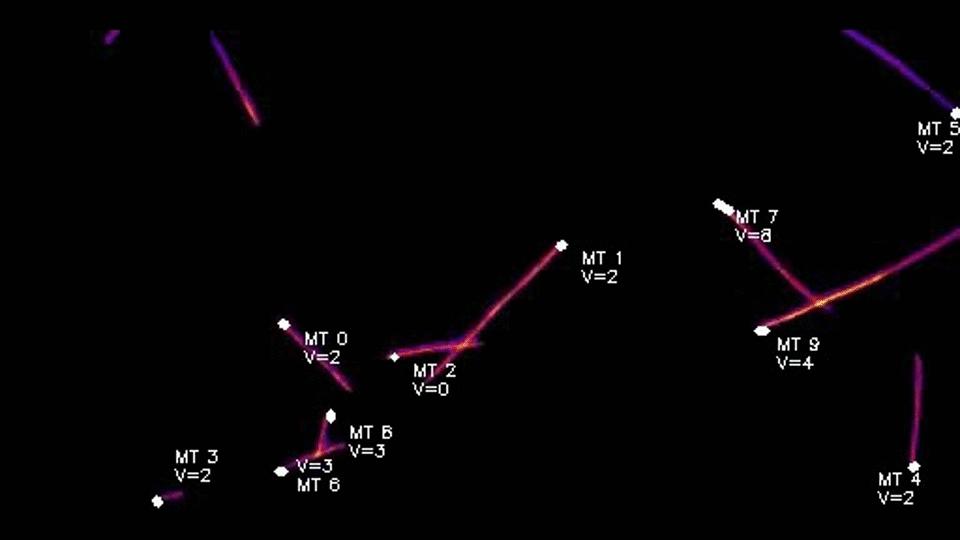}}
\\
\subfloat[]{\includegraphics[width=0.2\textwidth]{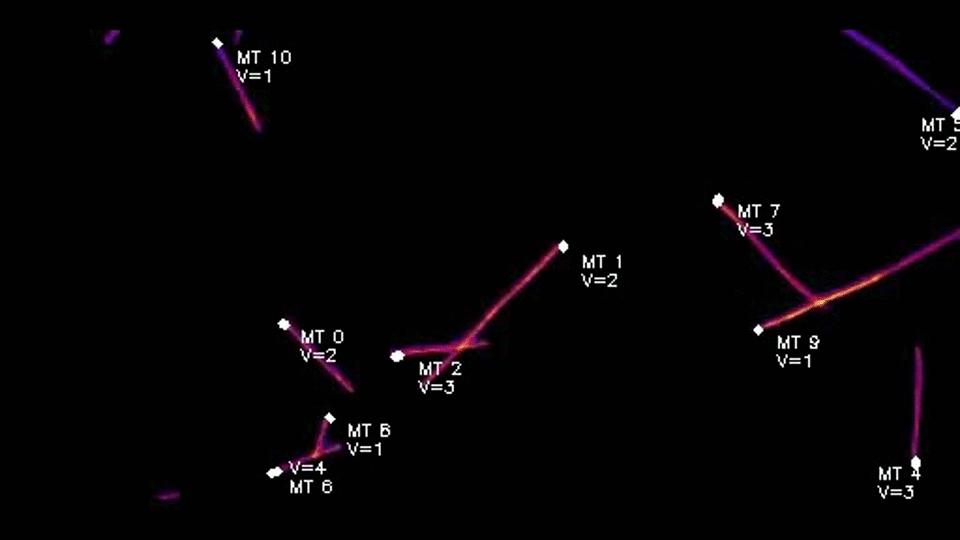}}
\hspace{.15cm}
\subfloat[]{\includegraphics[width=0.2\textwidth]{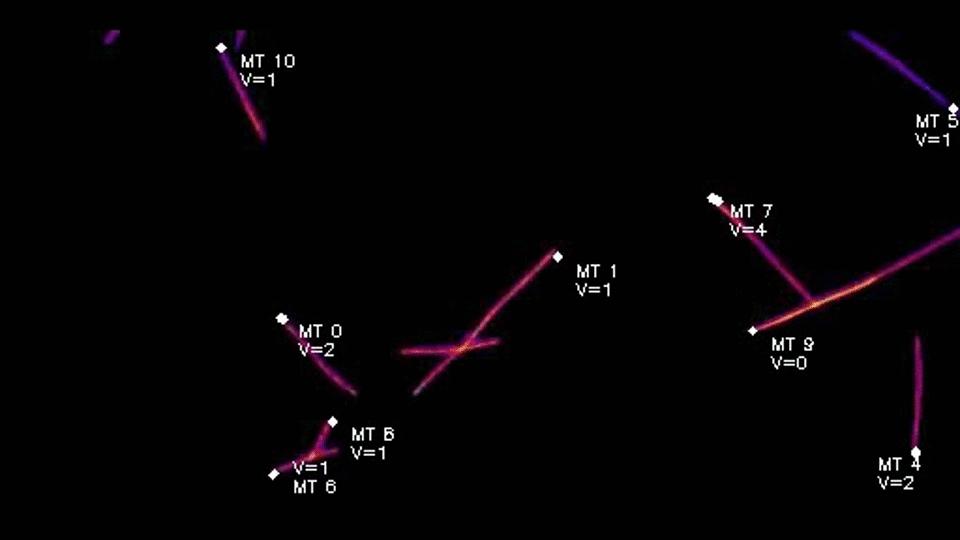}} 
\\
\subfloat[]{\includegraphics[width=0.2\textwidth]{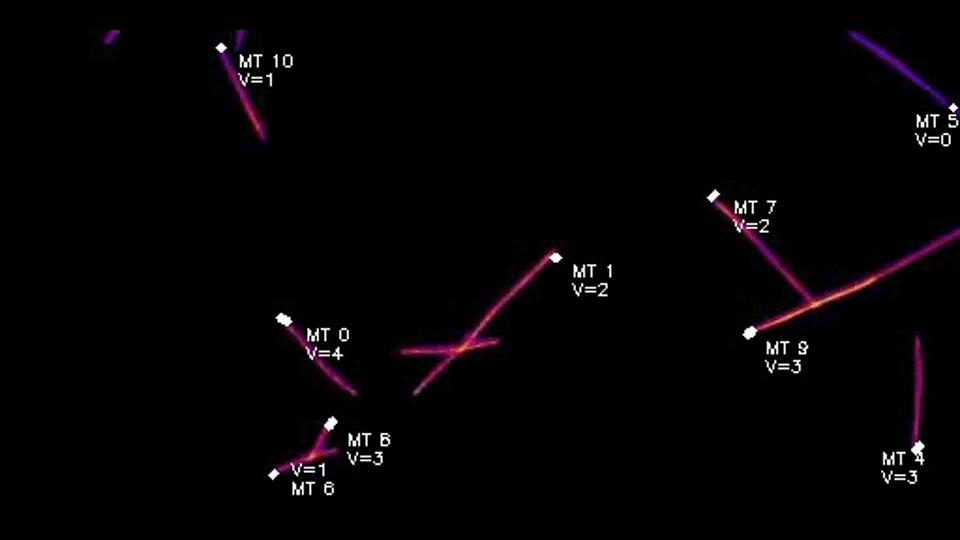}}
\hspace{.15cm}
\subfloat[]{\includegraphics[width=0.2\textwidth]{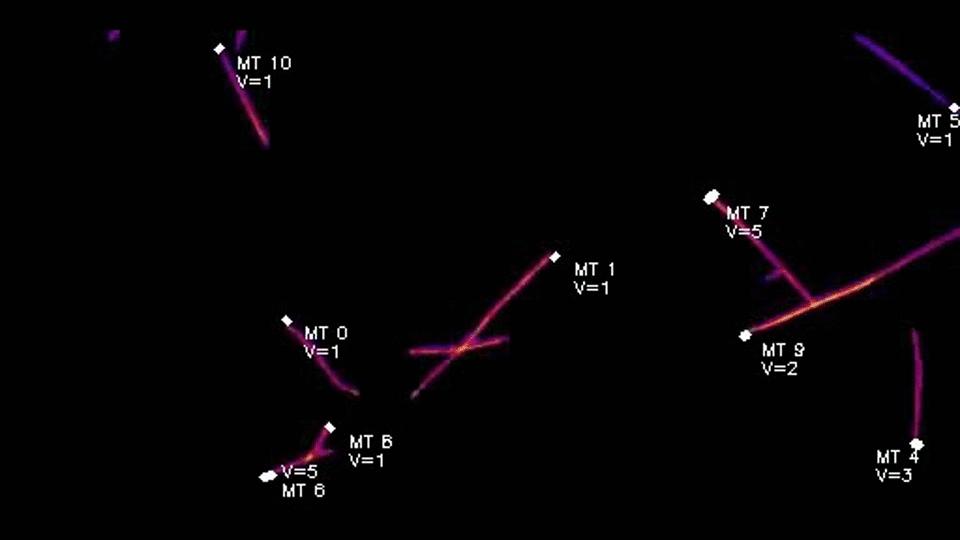}}
%
\caption{Resulted frames from segmentation and tracking of the respective time-lapse sequence with 4fps}
\label{Fig_app2}
\end{figure}

\textbf{Qualitative results for tracking.} Figures~\ref{Fig_app1} and~\ref{Fig_app2} respectively depict the input frames and the output frames which belong to a sample time-lapse sequence. Output frames represent the value of the velocity amplitude along the displacement for each individual MT.
 \label{FirstAppendix}
 
 \textbf{Definitions of quantitative measures.} In evaluating the segmentation, we define true positive as a count of the pixels that concurrently occur in both the segmented result and the ground truth. False positive, false negative and true negative are defined accordingly. 

We also define true positive in case of data association when the following conditions are fulfilled with respect to the obtained displacement and the ground truth vectors: 
\begin{itemize}
    \item The center of estimated displacement is less than 7 pixels away (Euclidean distance) form its corresponding ground truth value.
    \item Both vectors have less than 30 degrees angular difference.
    \item Both vectors have less than 10\% order of magnitude difference.
\end{itemize}%

\subsection{Acknowledgement}
We thank our colleague Paul Mooney (University of Wyoming) for providing the imaging data. We are also immensely grateful to John S. Oakey for his insight and expertise that greatly assisted the research. We thank Badrun Nessa Rahman and Yashasvi Bhat at the University of Central Florida for assistance with data labeling.

\bibliographystyle{ieeetr}
\bibliography{Ref.bib}

\end{document}